\pdfoutput=1

\documentclass{article} % For LaTeX2e

\usepackage{iclr2017_conference,times}
\usepackage[hidelinks]{hyperref}
\usepackage{url}

\usepackage[utf8]{inputenc} % allow utf-8 input
\usepackage[T1]{fontenc}    % use 8-bit T1 fonts
\usepackage{booktabs}       % professional-quality tables
\usepackage{amsfonts}       % blackboard math symbols
\usepackage{nicefrac}       % compact symbols for 1/2, etc.
\usepackage{microtype}      % microtypography
\usepackage{graphicx}
\usepackage{bm} 
\usepackage{amsmath}
\usepackage{amssymb}
\usepackage{multicol}
\usepackage{adjustbox}
\usepackage{subcaption}
\usepackage{multirow}
\usepackage{algorithm}
\usepackage[noend]{algpseudocode}
\def\BState{\State\hskip-\ALG@thistlm}

\allowdisplaybreaks[1]

\title{Learning and Policy Search in Stochastic\\ Dynamical Systems with Bayesian\\ Neural Networks}

\author{Stefan Depeweg\\
    Siemens AG and Technical University of Munich\\
  \texttt{stefan.depeweg@siemens.com} \\
    \And
    Jos\'e Miguel Hern\'andez-Lobato\\
    University of Cambridge\\
  \texttt{jmh233@cam.ac.uk} \\
    \And
    Finale Doshi-Velez\\
    Harvard University\\
  \texttt{finale@seas.harvard.edu} \\
    \And
    Steffen Udluft\\
    Siemens AG\\
  \texttt{steffen.udluft@siemens.com}
}
\iclrfinalcopy 
\begin{document}
\maketitle
\begin{abstract}
We present an algorithm for policy search in stochastic dynamical systems using
model-based reinforcement learning. The system dynamics are described with
Bayesian neural networks (BNNs) that include stochastic input variables.  These
input variables allow us to capture complex statistical
patterns in the transition dynamics (e.g. multi-modality and
heteroskedasticity), which are usually missed by alternative modeling approaches. After
learning the dynamics, our BNNs are then fed into an algorithm that performs
random roll-outs and uses stochastic optimization for policy learning. We train
our BNNs by minimizing $\alpha$-divergences with $\alpha = 0.5$, which usually produces better
results than other techniques such as variational Bayes. We illustrate the performance of our method by
solving a challenging problem where model-based approaches usually fail and by
obtaining promising results in real-world scenarios including the control of a
gas turbine and an industrial benchmark.
\end{abstract}
\vspace{-0.2cm}
\section{Introduction}

In model-based reinforcement learning, an agent uses its experience to
first learn a model of the environment and then uses that model to
reason about what action to take next. We consider the case in
which the agent observes the current state $\mathbf{s}_t$, takes some
action $\mathbf{a}$, and then observes the next state
$\mathbf{s}_{t+1}$. 
The problem of learning the model corresponds then
to learning 
%a transition function that determines the value of
%$\mathbf{s}_{t+1}$ as a function of $\mathbf{s}_t$ and
%$\mathbf{a}$. More generally, we desire 
a stochastic transition
function $p(\mathbf{s}_{t+1}|\mathbf{s}_t,\mathbf{a})$ specifying the
conditional distribution of $\mathbf{s}_{t+1}$ given $\mathbf{s}_t$
and $\mathbf{a}$.  Most classic control theory texts, e.g.
\cite{bertsekas1995dynamic}, will start with the most general model of
dynamical systems:
\begin{equation*}
\mathbf{s}_{t+1} = f( \mathbf{s}_t,\mathbf{a} , z ,  \mathcal{W} )
\end{equation*}
where $f$ is some deterministic function parameterized by weights
$\mathcal{W}$ that takes as input the current state $\mathbf{s}_t$,
the control signal $\mathbf{a}$, and some stochastic disturbance $z$.

However, to date, we have not been able to robustly learn dynamical
system models to such a level of generality.  Popular modes for
transition functions include Gaussian processes
\citep{rasmussen2003gaussian,ko2007gaussian,deisenroth2011pilco}, fixed
bases such as Laguerre functions \citep{wahlberg1991system}, and
adaptive basis functions or neural networks \citep{draeger1995model}.
All of these methods assume deterministic transition functions, perhaps
with some addition of Gaussian observation noise. Thus, they are
severely limited in the kinds of stochasticity---or transition
noise---they can express.  In many real-world scenarios
stochasticity may often arise due to some unobserved environmental
feature that can affect the dynamics in complex ways (such as
unmeasured gusts of wind on a boat).

In this work we use Bayesian neural networks (BNNs) in conjunction with a
random input noise source $z$ to express stochastic dynamics.  We take
advantage of a very recent inference advance based on $\alpha$-divergence
minimization \citep{hernandez2015black}, with $\alpha = 0.5$, to learn with high
accuracy BNN transition functions that are both scalable and expressive in
terms of stochastic patterns.
%---prior BNN approaches used variational methods
%\citep{Blundell2015,gal2015dropout}, probabilistic backpropagation
%\citep{hernandez2015probabilistic} and Markov Chain Monte Carlo methods
%\citep{balan2015bayesian,welling2011bayesian} for inference. 
Previous work 
achieved one but not both of these two characteristics. 
%We demonstrate that our combination of input noise and $\alpha$-divergence
%minimization allows us to model in a scalable way functions with complex
%stochastic patterns with high accuracy, something that was not previously
%possible in real-world reinforcement learning problems.

We focus our evaluation on the off-policy batch reinforcement learning
scenario, in which we are given an initial batch of data from an
already-running system and are asked to find a better (ideally
near-optimal) policy. Such scenarios are common in real-world industry
settings such as turbine control, where
exploration is restricted to avoid possible damage to the
system. We propose an algorithm that uses random roll-outs and
stochastic optimization for learning an optimal policy from the
predictions of BNNs. This method produces
(to our knowledge) the first model-based solution of a 20-year-old
benchmark problem: the Wet-Chicken
\citep{trespwc}. We also obtain very
promising results on a real-world application on controlling gas
turbines and on an industrial benchmark.

% FDV: I DON'T THINK THE PARAGRAPH BELOW IS PARTICULARLY RELEVANT 
% Significant progress has been made in applying neural network
% architectures to challenging problems (see \cite{schmidhuber2015deep}
% for an overview). Because they are trained by stochastic
% optimization\cite{welling2011bayesian} the result of training is only
% a maximum a posterior estimate over the distribution of all possible
% networks. Having a whole distribution and thereby being bayesian about
% neural networks would provide regularisation and, more importantly,
% uncertainty estimates over predictions.

\vspace{-0.1cm}

\def\i{^{(i)}}
\def\xopt{\vx_\star}
\def\xrec{\widetilde\vx}
\def\X{\calX}
\def\D{\calD}
\def\H{\mathrm{H}}
\newcommand{\x}{\mathbf{x}}
\newcommand{\EI}{\textrm{EI}}
\newcommand{\C}{\mathcal{C}}
\newcommand{\given}{\,|\,}
\newcommand{\DistGam}{\text{Gam}}

\section{Background}

\subsection{Model-Based Reinforcement Learning}

We consider  reinforcement learning problems in which an agent
acts in a stochastic environment by sequentially choosing actions over a
sequence of time steps, in order to minimize a cumulative cost. We assume
that our environment has some true dynamics
$T_\text{true}(\mathbf{s}_{t+1}|\mathbf{s},\mathbf{a})$, and we are given a cost function
$c(\mathbf{s}_t)$. In the model-based reinforcement learning setting, our goal
is to  learn an approximation
$T_\text{approx}(\mathbf{s}_{t+1}|\mathbf{s},\mathbf{a})$ for the true dynamics
based on collected samples
$(\mathbf{s}_t,\mathbf{a},\mathbf{s}_{t+1})$. The agent then tries to solve the 
control problem in which $T_\text{approx}$ is assumed to be the true dynamics.

\subsection{Bayesian neural networks with stochastic inputs}

Given data~${\mathcal{D} = \{ \mathbf{x}_n, \mathbf{y}_n \}_{n=1}^N}$, formed
by feature vectors~${\mathbf{x}_n \in \mathbb{R}^D}$ and targets~${\mathbf{y}_n
\in \mathbb{R}}^K$, we assume that~${\mathbf{y}_n =
f(\mathbf{x}_n,z_n;\mathcal{W}) + \bm \epsilon_n}$, where~$f(\cdot ,
\cdot;\mathcal{W})$ is the output of a neural network with weights
$\mathcal{W}$. The network receives as input the feature vector $\mathbf{x}_n$ and the
random disturbance $z_n \sim \mathcal{N}(0,\gamma)$.
The activation functions for the hidden layers are rectifiers:~${\varphi(x) = \max(x,0)}$.
The activation functions for the output layers are the identity function:~${\varphi(x) = x}$.
The network output is
corrupted by the additive noise variable~$\bm \epsilon_n \sim \mathcal{N}(\bm 0,\bm
\Sigma)$ with diagonal covariance matrix $\bm \Sigma$. The role of the noise disturbance $z_n$ is to
capture unobserved stochastic features that can affect the network's
output in complex ways. Without $z_n$, randomness is only given by the additive
Gaussian observation noise $\bm \epsilon_n$, which can only describe limited stochastic
patterns. 
The network has~$L$ layers, with~$V_l$ hidden units in layer~$l$,
and~${\mathcal{W} = \{ \mathbf{W}_l \}_{l=1}^L}$ is the collection of~${V_l
\times (V_{l-1}+1)}$ weight matrices.  
The $+1$ is introduced here to account
for the additional per-layer biases. 

One could argue why $\bm \epsilon_n$ is needed at all when we are
already using the more flexible stochastic model based on $z_n$. The reason for
this is that, in practice, we make predictions with the above model by averaging over a
finite number of samples of $z_n$ and $\mathcal{W}$.
By using $\bm \epsilon_n$, we obtain a predictive distribution whose density is well defined and given by a mixture of Gaussians.
If we eliminate $\bm \epsilon_n$, the predictive density is degenerate and given by a mixture of delta functions.

%The network has~$L$ layers, with~$V_l$ hidden units in layer~$l$,
%and~${\mathcal{W} = \{ \mathbf{W}_l \}_{l=1}^L}$ is the collection of~${V_l
%\times (V_{l-1}+1)}$ weight matrices.  
%The $+1$ is introduced here to account
%for the additional per-layer biases. 
%The outputs of the layers are~$\{
%\mathbf{z}_l \}_{l=0}^{L}$, where~$\mathbf{z}_0$ is the input layer
%and~$\mathbf{z}_L$ is the output layer.  The input to the $l$-th layer
%is~$\mathbf{a}_l = \mathbf{W}_l [ \mathbf{z}_{l-1} ; 1 ]$, where the operator
%$[ \,\cdot\, ; 1 \, ]$ concatenates an input vector with a coefficient for the
%bias weight.  

Let $\mathbf{Y}$ be an ${N\times K}$ matrix with the
targets~$\mathbf{y}_n$ and~$\mathbf{X}$ be an ${N\times D}$ matrix of
feature vectors~$\mathbf{x}_n$.
We denote by $\mathbf{z}$ the $N$-dimensional vector with the 
values of the random disturbances $z_1,\ldots,z_N$ that were used to generate the data.
The likelihood function is
\begin{align}
p(\mathbf{Y}\given\mathcal{W},\mathbf{z},\mathbf{X}) = 
\prod_{n=1}^N p(\mathbf{y}_n\given\mathcal{W},\mathbf{z},\mathbf{x}_n) =
\prod_{n=1}^N \prod_{k=1}^K \mathcal{N}(y_{n,k} \given f(\mathbf{x}_n,z_n;\mathcal{W}),\bm \Sigma)\,.\label{eq:likelihood}
\end{align}
The prior for each entry in $\mathbf{z}$ is $\mathcal{N}(0,\gamma)$.
We also specify a Gaussian prior distribution for each entry in each of the weight matrices in
$\mathcal{W}$. That is,
\begin{align}
p(\mathbf{z}) & =\prod_{n=1}^N \mathcal{N}(z_n|0,\gamma)\,, &
p(\mathcal{W}) &= \prod_{l=1}^L \prod_{i=1}^{V_l} \prod_{j=1}^{V_{l-1}+1} \mathcal{N}(w_{ij,l}\given0,\lambda)\,,
\label{eq:prior}
\end{align}
where $w_{ij,l}$ is the entry in the~$i$-th row and~$j$-th column
of~$\mathbf{W}_l$ and $\gamma$ and $\lambda$ are a prior variances.  The posterior
distribution for the weights~$\mathcal{W}$ and the random disturbances $\mathbf{z}$ is
given by Bayes' rule:
\begin{align}
p(\mathcal{W},\mathbf{z}\given\mathcal{D}) = \frac{p(\mathbf{Y}\given\mathcal{W},\mathbf{z},\mathbf{X})
p(\mathcal{W})p(\mathbf{z})}
{p(\mathbf{Y}\given\mathbf{X})}\,.
\end{align}
Given a new input vector~$\mathbf{x}_\star$, we can then
make predictions for $\mathbf{y}_\star$ using the
predictive distribution 
\begin{align}
p(\mathbf{y}_\star\given\mathbf{x}_\star,\mathcal{D}) &= \int\!\! 
\left[ \int 
\mathcal{N}(y_\star\given f(\mathbf{x}_\star,z_\star; \mathcal{W}), \bm \Sigma)\mathcal{N}(z_\star|0,1) \,dz_\star \right]
p(\mathcal{W},\mathbf{z}\given\mathcal{D})\, d\mathcal{W}\,d\mathbf{z}\,.\label{eq:predictive_distribution}
\end{align}
Unfortunately, the exact computation of (\ref{eq:predictive_distribution}) is intractable and we have to use approximations.

\vspace{-0.2cm}
 
\begin{figure}[t]
\centering
\includegraphics[width=0.8\textwidth]{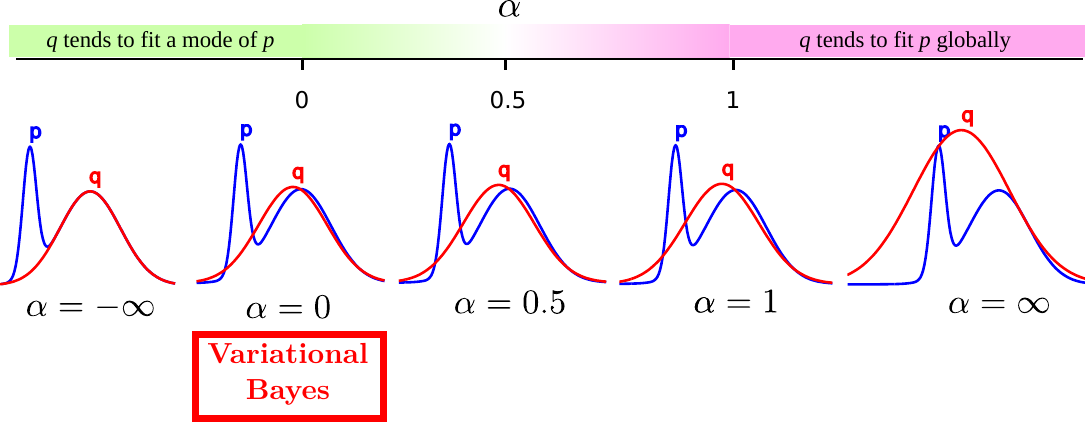}
\caption{Solution for the minimization of the $\alpha$-divergence between the
posterior $p$ (in blue) and the Gaussian approximation $q$ (in red and
unnormalized).  
%The solution of given by VB is obtained by setting $\alpha=0$.
%Values of $\alpha$ less than $0.5$ produce solutions that focus on fitting a
%local mode of $p$. Values of $\alpha$ larger than $0.5$ produce solutions that
%focus on fitting $p$ globally. 
Figure source \cite{minka2005divergence}. 
}\label{fig:alpha_divergence}
\end{figure}

\subsection{$\boldsymbol{\alpha}$-divergence minimization}

We approximate the exact posterior distribution $p(\mathcal{W},\mathbf{z}\given\mathcal{D})$
with the factorized Gaussian distribution
\begin{equation}
q(\mathcal{W},\mathbf{z}) = \left[ \prod_{l=1}^L\! \prod_{i=1}^{V_l}\! 
\prod_{j=1}^{V_{l\!-\!1}\!+\!1} \mathcal{N}(w_{ij,l}| m^w_{ij,l},v^w_{ij,l})\right]
\left[\prod_{n=1}^N \mathcal{N}(z_n \given m_n^z, v_n^z) \right]\,.\label{eq:posterior_approximation}
\end{equation}
The parameters~$m^w_{ij,l}$,~$v^w_{ij,l}$ and
~$m^z_n$,~$v^z_n$ are determined by
minimizing a divergence between 
$p(\mathcal{W},\mathbf{z}\given\mathcal{D})$
and the approximation $q$.
After fitting $q$, we make predictions by replacing $p(\mathcal{W},\mathbf{z}\given\mathcal{D})$ 
with $q$ in (\ref{eq:predictive_distribution})
and approximating the integrals in (\ref{eq:predictive_distribution})
with empirical averages over samples of $\mathcal{W}\sim q$.

We aim to adjust the parameters of (\ref{eq:posterior_approximation}) by minimizing the
$\alpha$-divergence between $p(\mathcal{W},\mathbf{z}\given\mathcal{D})$ and $q(\mathcal{W},\mathbf{z})$ \citep{minka2005divergence}:
\begin{equation}
\text{D}_\alpha[p(\mathcal{W},\mathbf{z}\given\mathcal{D})||q(\mathcal{W},\mathbf{z})] = \frac{1}{\alpha(\alpha-1)} \left( 1 - \int p(\mathcal{W},\mathbf{z}\given \mathcal{D})^\alpha 
q(\mathcal{W},\mathbf{z})^{(1 - \alpha)}\right)\,d\mathcal{W}\,d\mathbf{z}\,,
\label{eq:alpha_divergence}
\end{equation}
which includes a parameter $\alpha\in \mathbb{R}$ that controls the properties
of the optimal $q$. Figure \ref{fig:alpha_divergence} illustrates these properties 
for the one-dimensional case. When $\alpha \geq 1$, $q$ tends to cover the whole posterior distribution $p$.
When $\alpha \leq 0$, $q$ tends to fit a local mode in $p$.
The value $\alpha=0.5$ is expected to achieve a balance between these two tendencies.
Importantly,
when $\alpha\rightarrow 0$, the solution obtained is the same as with variational
Bayes (VB) \citep{wainwright2008graphical}.

%, which minimizes the energy function 
%\begin{align}
%E_\text{VB}(q) = -\sum_{n=1}^N \mathbf{E}_q \left[ \log p(\mathbf{y}_n | \mathbf{x}_n ,\mathcal{W},\bm \Gamma)\right] +
%\text{KL}[q||p(\mathcal{W}|\lambda)]\,,\label{eq:VB}
%\end{align}
%where $\text{KL}[q||p(\mathcal{W}|\lambda)]$ denotes the Kullback-Leibler
%divergence between $q$ and the prior (\ref{eq:prior}). 

The direct minimization of (\ref{eq:alpha_divergence}) is infeasible in practice for arbitrary $\alpha$.
Instead, we follow \cite{hernandez2015black} and optimize an energy function
whose minimizer corresponds to a local minimization of $\alpha$-divergences, with one
$\alpha$-divergence for each of the $N$ likelihood factors in
(\ref{eq:likelihood}). 
Since $q$ is Gaussian and
the priors $p(\mathcal{W})$ and $p(\mathbf{z})$ are also Gaussian, we represent $q$ as 
\begin{align}
q(\mathcal{W},\mathbf{z}) \propto \left[ \prod_{n=1}^N f(\mathcal{W}) f_n(z_n) \right] p(\mathcal{W})p(\mathbf{z})\,,
\end{align}
where $f(\mathcal{W})$ is a Gaussian factor
that approximates the geometric mean of the $N$ likelihood factors in
(\ref{eq:likelihood}) as a function of $\mathcal{W}$.
Each $f_n(z_n)$ is also a Gaussian factor that approximates
the $n$-th likelihood factor in
(\ref{eq:likelihood}) as a function of $z_n$.
We adjust $f(\mathcal{W})$ and the $f_n(z_n)$ by minimizing
local $\alpha$-divergences.
In particular, we minimize the energy function
\begin{align}
E_\alpha(q) = -\log Z_q - \frac{1}{\alpha} \sum_{n=1}^N
\log \mathbf{E}_{\mathcal{W},z_n\sim\, q}\left[ \left( \frac{p(\mathbf{y}_n \given \mathcal{W}, \mathbf{x}_n, z_n, \bm \Sigma)}
{f(\mathcal{W})f_n(z_n)}\right)^\alpha \right]\,,\label{eq:energy}
\end{align}
\citep{hernandez2015black}, where $f(\mathcal{W})$ and $f_n(z_n)$ are in exponential Gaussian form and parameterized in terms
of the parameters of $q$ and the priors $p(\mathcal{W})$ and $p(z_n)$, that is,
\begin{align}
f(\mathcal{W}) =& \exp \left\{ \sum_{l=1}^L\! \sum_{i=1}^{V_l}\! 
\sum_{j=1}^{V_{l\!-\!1}\!+\!1} \frac{1}{N}\left( \frac{\lambda v^w_{i,j,l}}{\lambda - v^w_{i,j,l}} w_{i,j,l}^2 + 
\frac{m^w_{i,j,l}}{v^w_{i,j,l}} w_{i,j,l} 
\right) \right\} \propto \left[\frac{q(\mathcal{W})}{p(\mathcal{W})}\right]^{\frac{1}{N}}\,,\\
f_n(z_n) =& \exp \left\{ \frac{ \gamma v^z_n}{\gamma - v^z_n} z_n^2 + 
\frac{m^z_n}{v^z_n} z_n \right\} \propto \frac{q(z_n)}{p(z_n)}\,,
\end{align}
and $\log Z_q$ is the logarithm of the normalization constant of the exponential Gaussian form of $q$:
\begin{align}
\log Z_q = \sum_{l=1}^L\! \sum_{i=1}^{V_l}\! 
\sum_{j=1}^{V_{l\!-\!1}\!+\!1} \left[ \frac{1}{2} \log \left( 2 \pi v^w_{i,j,l} \right) + \frac{\left(m^w_{i,j,l}\right)^2}{v^w_{i,j,l}} \right] + 
\sum_{n=1}^N \left[ \frac{1}{2} \log \left( 2 \pi v^z_n \right) + \frac{\left(m^z_n\right)^2}{v^z_n} \right]
\,.\label{eq:logZq}
\end{align}
The scalable optimization of (\ref{eq:energy}) is done in practice by using
stochastic gradient descent. For this, we subsample the sums for $n=1,\ldots,N$
in (\ref{eq:energy}) and (\ref{eq:logZq}) using mini-batches and approximate
the expectations over $q$ in (\ref{eq:energy}) with an average over $K$ samples
drawn from $q$. We can then use the reparametrization trick
\citep{kingma2015variational} to obtain gradients from the resulting stochastic
approximator to (\ref{eq:energy}). The hyper-parameters $\bm \Sigma$, $\lambda$
and $\gamma$ can also be tuned by minimizing (\ref{eq:energy}). In practice we
only tune $\bm \Sigma$ and keep $\lambda = 1$ and $\gamma = d$. The latter means
that the prior scale of each $z_n$ grows with the data dimensionality. This
guarantees that, a priori, the effect of each $z_n$ in the neural network's output
does not diminish when more and more features are available.

Minimizing (\ref{eq:energy}) when $\alpha \rightarrow 0$ is equivalent to
running the method VB \citep{hernandez2015black}, which has recently been used
to train Bayesian neural networks in reinforcement learning problems
\citep{Blundell2015,Houthooft2016,Gal2016}. However, we propose to minimize
(\ref{eq:energy}) using $\alpha = 0.5$, which often results in 
better test log-likelihood values.

% than VB and than $\alpha = 1.0$.

%  \citep{hernandez2015black}.
%We expect the improved predictive performance of $\alpha = 0.5$ to result also
%in better performing policies.

We have also observed $\alpha = 0.5$ to be more robust than VB when
$q(\mathbf{z})$ is not fully optimized. In particular, $\alpha = 0.5$ can
still capture complex stochastic patterns even when we do not learn $q(\mathbf{z})$
and instead keep it fixed to the prior $p(\mathbf{z})$. By contrast, VB fails
completely in this case (see Appendix \ref{sec:toy_problems}). 

%Another problem of VB is that it can be
%significantly affected by overfitting problems. For example, with limited
%data, there are many spurious values of $\mathcal{W}$ and $\mathbf{z}$ that 
%produce very small training errors. Each of these spurious solutions
%represents a mode in the posterior with poor generalization
%performance and VB is likely to fit just one of these modes. By contrast, $\alpha = 0.5$ moves away from the mode
%seeking properties of VB, resulting in a more robust predictive performance
%with limited data. In our experiments, we also observe that $\alpha =
%0.5$ is usually better than $\alpha = 1.0$. This latter technique has
%an excessive tendency to cover with $q$ several distant posterior modes, which
%can result in poor predictive performance.

\vspace{-0.2cm}
\section{Policy search using BNNs with stochastic inputs}\label{sec:policy_search}

%using $\boldmath{\alpha}$-divergence minimization}

%<<<<<<< HEAD
%In the previous section, we described how to to learn a transition
%function $T(s'|s,a)$ using Bayesian networks with
%$\boldmath{\alpha}$-divergence minimization.  We now describe how to
%find an optimal policy for that model.  Suppose that we have a policy
%class that is parameterized by the weights $\mathcal{W}_{\pi}$ (in the
%following we will use FILL as the policy class).  The expected cost of
%a policy $\pi$ is obtained by averaging over multiple roll-outs: Given
%a starting state $\mathbf{s}_0$, we simulate the evolution of the
%system over a fixed horizon $T$ using the probabilistic predictions of
%the Bayesian network and the actions produced by the current policy.
%This procedure allows us to evaluate the performance for any
%particular cost function. If model, policy and cost function are
%differentiable, we are then able to tune $\mathcal{W}_{\pi}$ by
%gradient descent over the roll-out average.  CITE 
%=======

% EMPHASIZE THAT THIS IS WHERE THE NOVEL/NEW RESEARCH STARTS.  ALSO,
% PILCO CAN'T BE THE CANONICAL REFERENCE FOR THIS APPROACH, CAN IT?
% WHOM DO THEY SITE? 

We now describe a gradient-based policy search algorithm that uses the
BNNs with stochastic disturbances from the previous section.  
%While
%gradient-based methods are the mainstay of policy search algorithms,
%we demonstrate how using a BNN allows us to learn near-optimal
%policies from only observational data.  
The motivation for our
approach lies in its applicability to industrial systems: we wish to
estimate a policy in parametric form, using only an available batch of
state transitions obtained from an already-running system. We assume
that the true dynamics present stochastic patterns that arise due to
some unobserved process affecting the system in complex ways.

Model-based policy search methods include two key parts
\citep{deisenroth2013survey}. The first part consists in learning a
dynamics model from data in the form of state transitions
$(\mathbf{s}_t,\mathbf{a}_t,\mathbf{s}_{t+1})$, where $\mathbf{s}_t$
denotes the current state, $\mathbf{a}_t$ is the action applied and
$\mathbf{s}_{t+1}$ is the resulting state. The second part consists in
learning the parameters $\mathcal{W}_\pi$ of a deterministic policy
function $\pi$ that returns the optimal action $\mathbf{a}_t =
\pi(\mathbf{s}_t; \mathcal{W}_\pi)$ as function of the current state
$\mathbf{s}_t$.  The policy function can be a neural
network with deterministic weights given by $\mathcal{W}_\pi$.

%This process can be done either jointly or as a two-stage
%process. In this work we will use a two-stage approach. 

The first part in the aforementioned procedure is a standard regression task,
which we solve by using the modeling approach from the previous section. We
assume the dynamics to be stochastic with the following true transition model:
\begin{equation}\label{eq:transitions}
\mathbf{s}_t = f_\text{true}(\mathbf{s}_{t-1},\mathbf{a}_{t-1},z_t;\mathcal{W}_\text{true})\,,\quad z_t \sim \mathcal{N}(0,\gamma)\,. 
\end{equation} 
where the input disturbances $z_t \sim \mathcal{N}(0,\gamma)$
account for the stochasticity in the dynamics. When the Markov state $\mathbf{s}_t$ is hidden 
and we are given only observations $\mathbf{o}_t$ , we can use the time embedding theorem 
using a suitable window of length $n$ and approximate:
\begin{equation}\label{time_emb}
\hat{\mathbf{s}}(t) = [\mathbf{o}_{t-n},\cdots,\mathbf{o}_t]\,.
\end{equation}
The transition model in equation \ref{eq:transitions} specifies a probability distribution $p(\mathbf{s}_t|\mathbf{s}_{t-1},\mathbf{a}_{t-1})$
that we approximate using a BNN with stochastic inputs:
\begin{equation}\label{eq:trans_appr}
p(\mathbf{s}_t|\mathbf{s}_{t-1},\mathbf{a}_{t-1}) 
\approx \int \mathcal{N}(\mathbf{s}_{t}|f(\mathbf{s}_{t-1},\mathbf{a}_{t-1},z_t; \mathcal{W}),\bm \Sigma)  
q(\mathcal{W}) \mathcal{N}(z_t|0,\gamma)\,d\mathcal{W}\,dz_t\,,
\end{equation}
where the feature vectors in our BNN are now $\mathbf{s}_{t-1}$ and $\mathbf{a}_{t-1}$ and
the targets are given by $\mathbf{s}_{t}$.  
In this expression, the integration with respect to $\mathcal{W}$ accounts for
stochasticity arising from lack of knowledge of the model parameters,
while the integration with respect to $z_t$ accounts for
stochasticity arising from unobserved processes that cannot be modeled.
In practice, these integrals are approximated by an average over samples
of $z_t \sim \mathcal{N}(0,\gamma)$ and $\mathcal{W}\sim q$.

In the second part of our model-based policy search algorithm,
we optimize the parameters $\mathcal{W}_{\pi}$ of a policy that minimizes the sum of expected cost over a finite horizon $T$ with respect
to our belief $q(\mathcal{W})$. This expected cost is obtained by averaging
over multiple virtual roll-outs. 
For each roll-out we sample $\mathcal{W}_i\sim q$ and then simulate state trajectories using 
the model $\mathbf{s}_{t+1}=f(\mathbf{s}_t,\mathbf{a}_t,z_t;\mathcal{W}_i)+\bm \epsilon_{t+1}$ with 
policy $\mathbf{a}_t = \pi(\mathbf{s}_t;\mathcal{W}_\pi)$,
input noise $z_t \sim \mathcal{N}(0,\gamma)$ and additive noise $\bm \epsilon_{t+1} \sim \mathcal{N}(\bm 0, \bm \Sigma)$.
This procedure allows us to obtain estimates of the policy's expected cost for any particular
cost function. If model, policy and cost function are differentiable, we are
then able to tune $\mathcal{W}_{\pi}$ by stochastic gradient descent over the roll-out average.

\begin{figure}
\resizebox{0.525\textwidth}{!}{
\adjustbox{valign=t}{\begin{minipage}[t]{.525\textwidth}
\vspace{-0.3cm}
\begin{algorithm}[H]
\begin{algorithmic}[1]
\State {\bf Input:} $\mathcal{D} = \{\mathbf{s}_n,a_n,\bm \Delta_n\}$ for $n \in 1..N$
\State \textit{Fit} $q(\mathcal{W})$ \textit{and} $\bm \Sigma$ \textit{by optimizing} (\ref{eq:energy}).
\Function{unfold}{$\mathbf{s}_0$}
	\State $\textit{sample} \{\mathcal{W}^{1},..,\mathcal{W}^{K}\} \textit{ from } q(\mathcal{W})$
	\State $C \leftarrow 0$
	\For{$k = 1:K$}
		\For{$t = 0:T$}
			\State $z_{t+1}^k \sim \mathcal{N}(0,\gamma)$
			\State $\bm \Delta_t \leftarrow f(\mathbf{s}_t, \pi(\mathbf{s}_t; \mathcal{W}_\pi),z_{t+1}^k ; \mathcal{W}^k)$
			\State $\bm \epsilon_{t+1}^k \sim \mathcal{N}(\mathbf{0},\bm \Sigma)$
			\State $\mathbf{s}_{t+1} \leftarrow \mathbf{s}_t + \bm \Delta_t + \bm \epsilon_{t+1}^k$
			\State $C \leftarrow C + c(\mathbf{s}_{t+1})$
		\EndFor
	\EndFor
	\State \Return $C / K$
\EndFunction
\State  \textit{Fit} $\mathcal{W}_{\pi}$ \textit{by optimizing} $\frac{1}{N} \sum_{n=1}^N$ \Call{unfold}{$\mathbf{s}_n$}
\end{algorithmic}
\caption{Model-based policy search using\\ Bayesian neural networks with stochastic inputs.}
\label{algo2}
\end{algorithm}
\end{minipage}
}}
\adjustbox{valign=t}{
\begin{minipage}[t]{0.47\textwidth}
\centering
\begin{subfigure}{.48\textwidth}
\refstepcounter{subfigure}\label{fig:wc11}
\centering
\includegraphics[width=\textwidth]{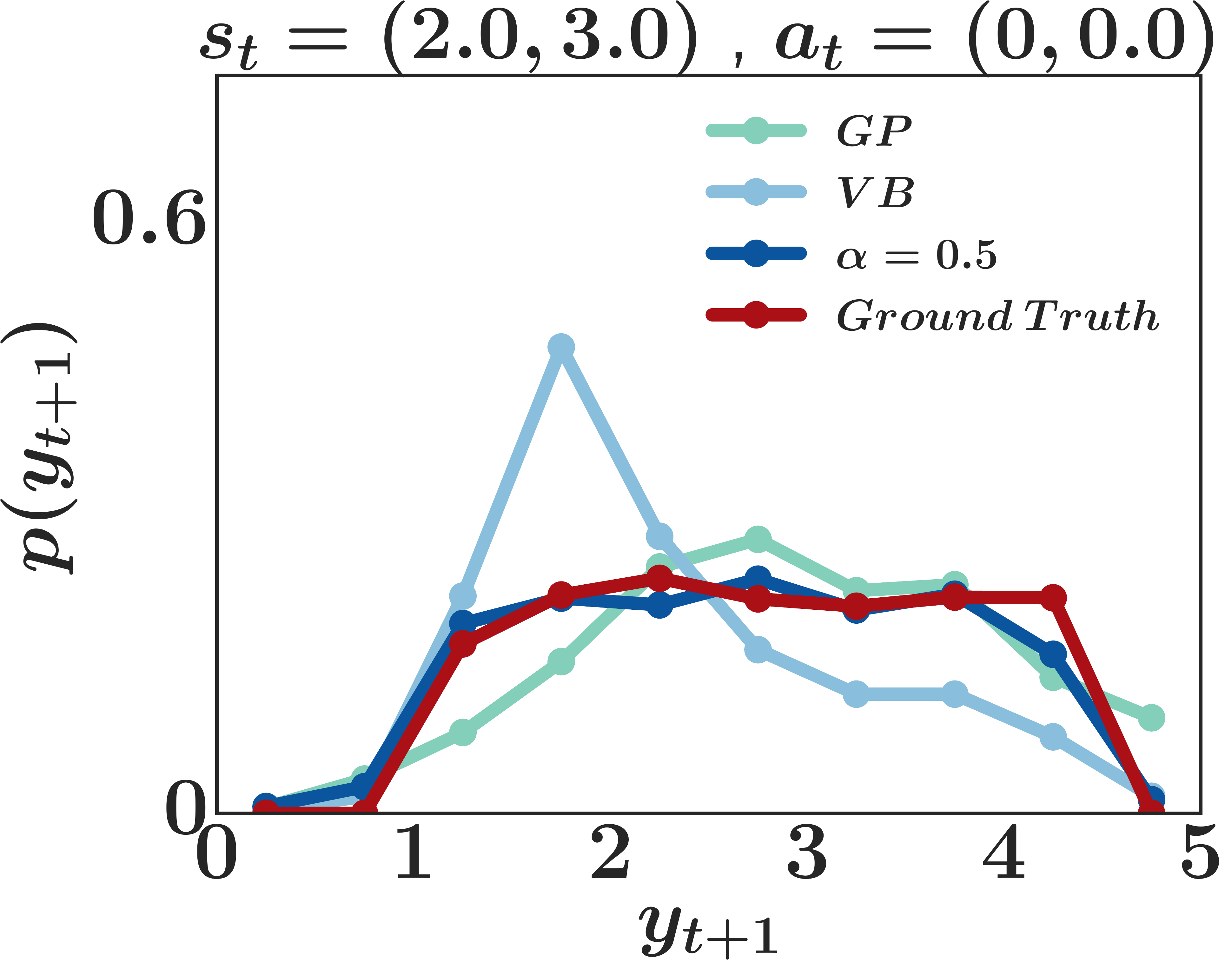}\\
%\hspace{0.65cm}(a)
%\label{fig:wc11}
\end{subfigure}
\begin{subfigure}{.48\textwidth}
\centering
\includegraphics[width=\textwidth]{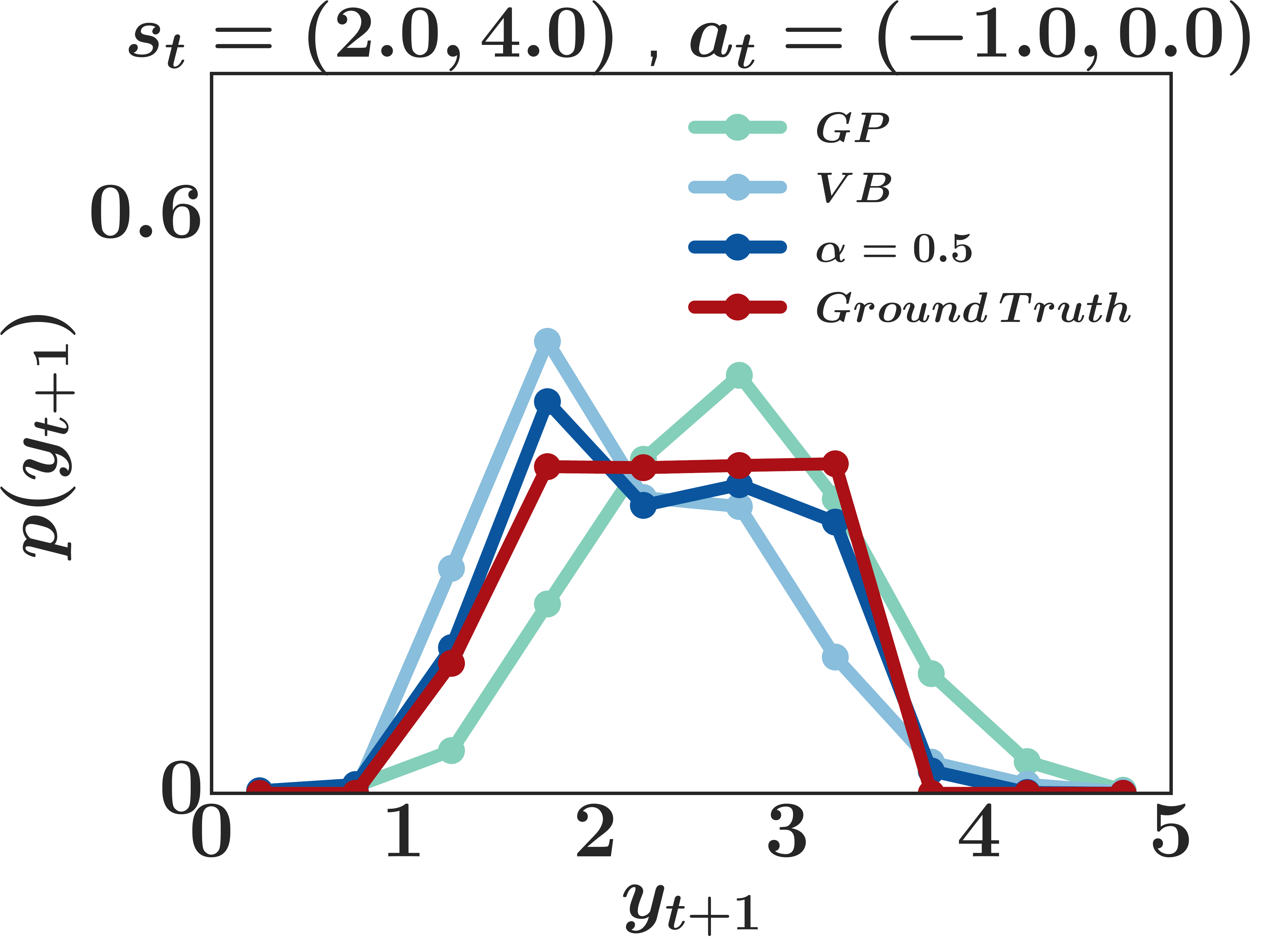}\\
%\hspace{0.575cm}(b)
\refstepcounter{subfigure}\label{fig:wc12}
\end{subfigure}
\begin{subfigure}{.48\textwidth}
\centering
\includegraphics[width=\textwidth]{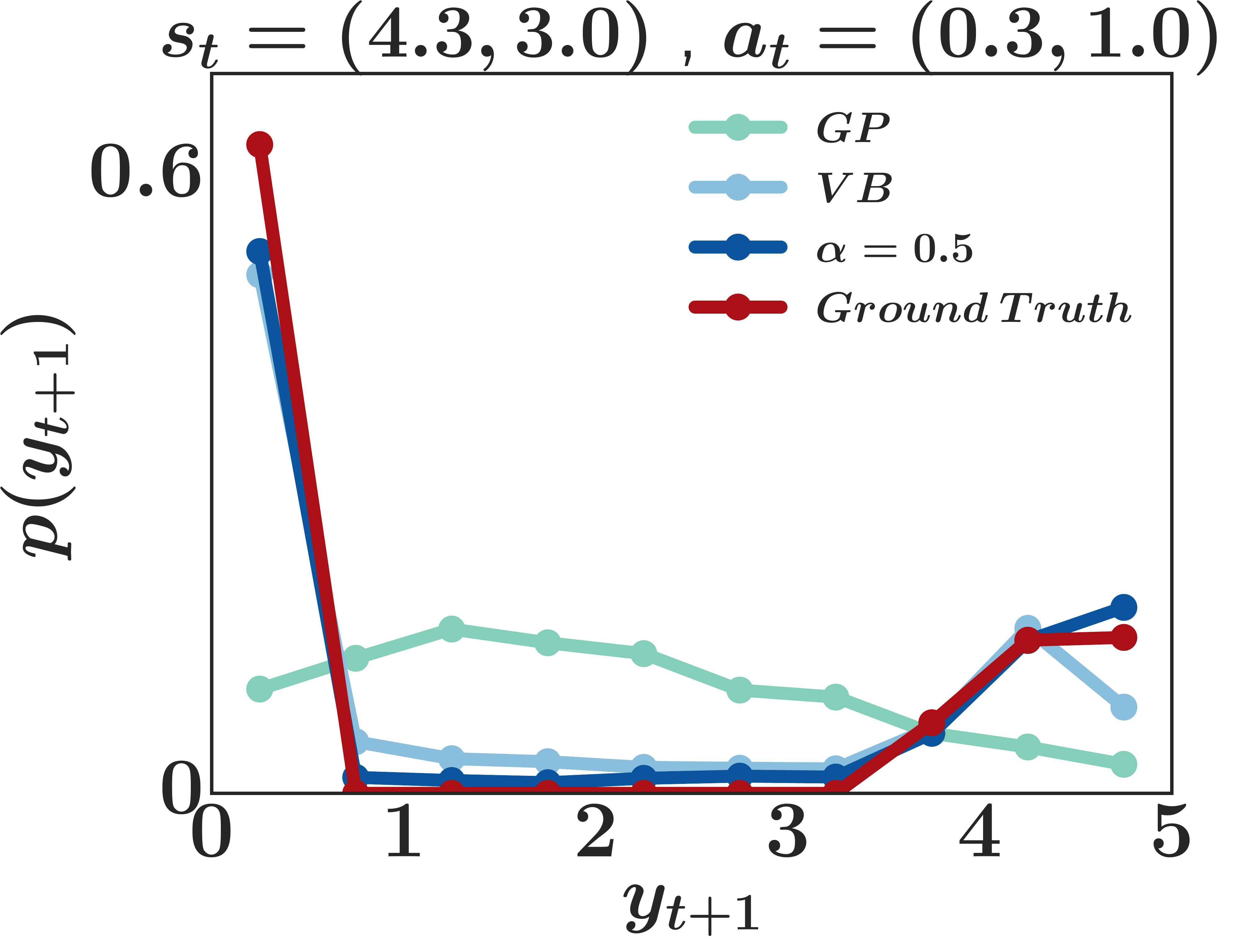}\\
%\hspace{0.55cm}(c)
%\refstepcounter{subfigure}\
\refstepcounter{subfigure}\label{fig:wc13}
\end{subfigure}
\begin{subfigure}{.48\textwidth}
\centering
\includegraphics[width=\textwidth]{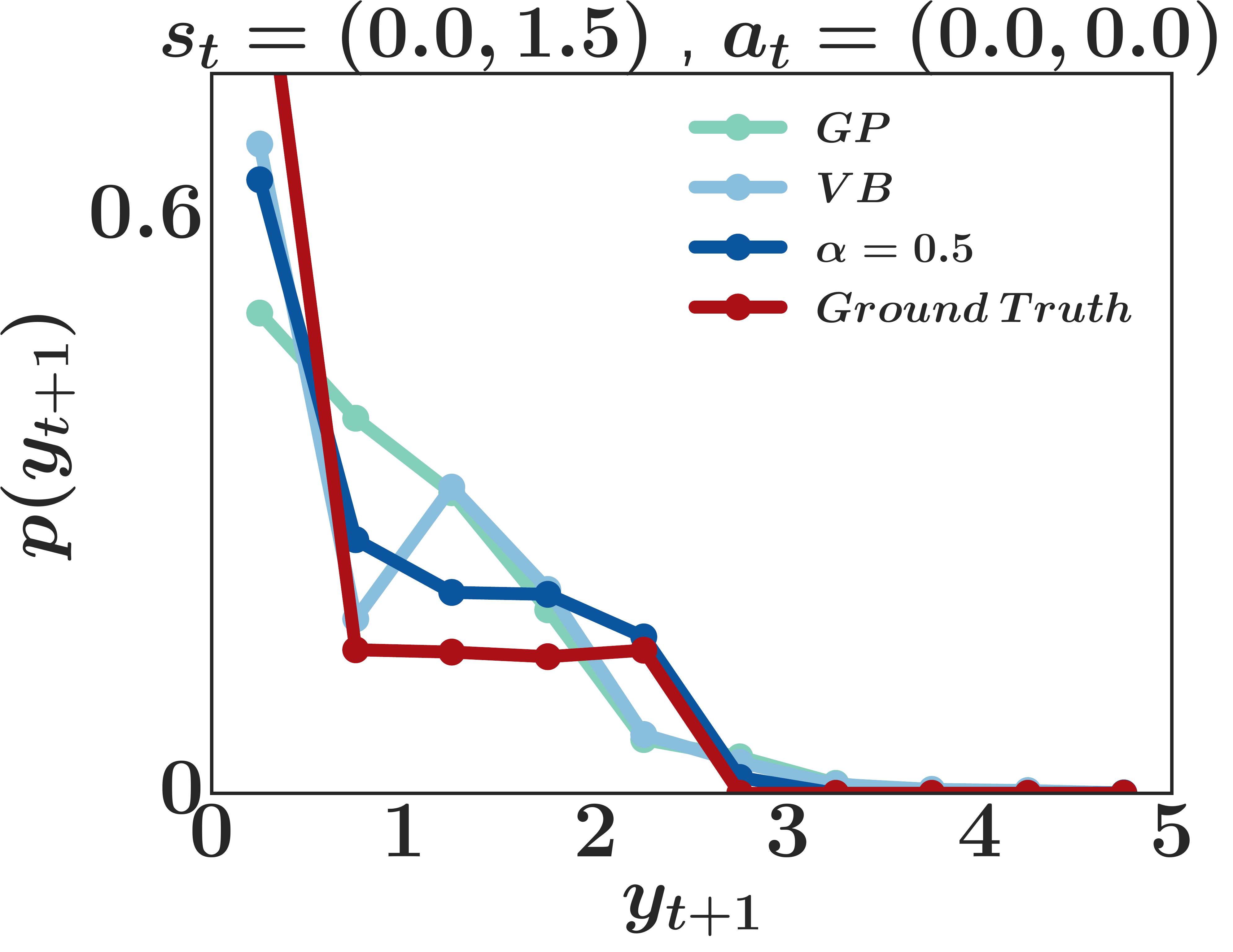}\\
%\hspace{0.50cm}(d)
%\refstepcounter{subfigure}
\refstepcounter{subfigure}\label{fig:wc14}
\end{subfigure}
\caption{Predictive distribution of $y_t$ given by different methods
in four different scenarios. Ground truth (red) is obtained by
sampling from the real dynamics.}
\label{fig:wc_model}
\vspace{-0.2cm}
\end{minipage}}
\end{figure}
Given a cost function $c(\mathbf{s}_t)$, the objective to be optimized by our policy search algorithm is
\begin{align}
\textstyle J(\mathcal{W}_{\pi}) & = \textstyle \mathbf{E}\left[\sum_{t=1}^{T}c(\mathbf{s}_{t})\right]\,.\label{eq:exact_cost}
\end{align}
We approximate (\ref{eq:exact_cost}) by using (\ref{eq:trans_appr}), replacing $\mathbf{a}_t$ with $\pi(\mathbf{s}_t;
\mathcal{W}_\pi)$ and using sampling to approximate the expectations:
\vspace{-0.1cm}
\begin{small}
\begin{align}
J(\mathcal{W}_{\pi}) = & \int\left[\sum_{t=1}^{T}c(\mathbf{s}_{t})\right]\left[\prod_{t=1}^{T}\int\mathcal{N}(\mathbf{s}_{t}|
        f(\mathbf{s}_{t-1},\pi(\mathbf{s}_{t-1};\mathcal{W}_\pi),z_t;
        \mathcal{W}),\bm{\Sigma})q(\mathcal{W})\mathcal{N}(z_t|0,\gamma)\,d\mathcal{W}\,dz_t\right]\nonumber\\
         &p(\mathbf{s}_0)d\mathbf{s}_{0}\cdots d\mathbf{s}_{T}\nonumber\\
     = & \int\left[\sum_{t=1}^{T}c(\mathbf{s}_{t}^{\mathcal{W},\{z_{1},
        \ldots,z_{t}\},\{\bm{\epsilon}_{1},\ldots,\bm{\epsilon}_{t}\},
        \mathcal{W}_\pi})\right]
        q(\mathcal{W})d\mathcal{W}\left[\prod_{t=1}^{T}\mathcal{N}(\bm{\epsilon}_{t}|\mathbf{0},\bm{\Sigma})\mathcal{N}(z_{t}|0,\gamma)
        d\bm{\epsilon}_{t}dz_t\right] p(\mathbf{s}_0)\,d\mathbf{s}_0\nonumber\\
    \approx & \frac{1}{K}\sum_{k=1}^{K} \left[\sum_{t=1}^{T}c(\mathbf{s}_{t}^{\mathcal{W}^{k},\{z_{1}^{k},\ldots,z_{t}^{k}\},
        \{\bm{\epsilon}_{1}^{k},\ldots,\bm{\epsilon}_{t}^{k}\},\mathcal{W}_\pi})\right]\,.\label{eq:approximation_cost}
\end{align}
\end{small}The first line in (\ref{eq:approximation_cost}) is obtained by using the assumption that the dynamics
are Markovian with respect to the current state and the current action and by replacing
$p(\mathbf{s}_t|\mathbf{s}_{t-1},\mathbf{a}_{t-1})$ with the right-hand side of (\ref{eq:trans_appr}).
In the second line,
$\mathbf{s}_{t}^{\mathcal{W},\{z_{1},
        \ldots,z_{t}\},\{\bm{\epsilon}_{1},\ldots,\bm{\epsilon}_{t}\},
        \mathcal{W}_\pi}$ 
         is the state that is obtained at time $t$ in a roll-out
generated by using a policy with parameters $\mathcal{W}_\pi$, a transition function parameterized by
$\mathcal{W}$ and input noise $z_1,\ldots,z_{t}$, with additive noise values
$\bm{\epsilon}_{1},\ldots,\bm{\epsilon}_{t}$.
In the last line we have approximated the integration with respect to
$\mathcal{W},z_1,\ldots,z_T$, $\bm{\epsilon}_1,\ldots,\bm{\epsilon}_T$
and $\mathbf{s}_0$ by averaging over $K$ samples of these variables. To
sample $\mathbf{s}_0$, we draw this variable uniformly from the available transitions $(\mathbf{s}_t,\mathbf{a}_t,\mathbf{s}_{t+1})$.

The expected cost (\ref{eq:exact_cost}) can then be optimized by stochastic
gradient descent using the gradients of the Monte Carlo approximation given by the last line of
(\ref{eq:approximation_cost}). Algorithm \ref{algo2} computes this Monte Carlo
approximation. The gradients can then be obtained using automatic
differentiation tools such as Theano \citep{2016arXiv160502688short}. 
Note that Algorithm \ref{algo2} uses the BNNs to make predictions for the change in the state
$\Delta_t = \mathbf{s}_{t+1} - \mathbf{s}_{t}$ instead of for the next state
$\mathbf{s}_{t+1}$ since this approach often performs better in practice \citep{deisenroth2011pilco}.

%at time $t$ we sample $\mathcal{W}_t^k$ from $q(\mathcal{W})$ and $\bm
%\epsilon_t$ from $\mathcal{N}(0,\bm \Gamma)$ and apply the policy with
%parameters $\mathcal{W}_\pi$ to the current state. 

%Because we are assuming a stochastic system,
%this sampling process is a model of the underlying randomness in the stochastic
%process. The last equation is a Monte-Carlo estimate of the integral. This now
%allows  us to calculate the gradient $\nabla_{\mathcal{W}_\pi}J$
%for optimizing the policy.
 
%We now describe an algorithm to compute the Monte Carlo approximation used in
%(\ref{eq:approximation}). However, instead of sampling a new $\mathcal{W}^k_t$
%at each roll-out step, we only sample $K$ different values for
%$\mathbf{\mathcal{W}}$ at the beginning of the roll-out and then choose
%uniformly at random one of these samples at each roll-out step.
%This makes more efficient the implementation of the algorithm 
%in automatic differentiation environments such as theano.

% at
%random. The resulting algorithm is given in \ref{algo2}. Also note that the
%outer loop over $K$ can be done in parallel  because each roll-out is
%independent of each other.  

\vspace{-0.15cm}
\section{Experiments}

We now evaluate the performance of our algorithm for policy search in different
benchmark problems. These problems are chosen based on two reasons. First, they
contain complex stochastic dynamics and second, they represent real-world
applications common in industrial settings. A theano implementation
of algorithm \ref{algo2}  is available online\footnote{ \url{https://github.com/siemens/policy_search_bb-alpha}}.
%
%Our goal is to empirically support the following claims:
%\begin{enumerate}
%\item BNNs with stochastic inputs and trained by minimizing $\alpha$-divergences
%can successfully learn complex stochastic patterns in the system dynamics.
%\item Algorithm \ref{algo2} can exploit the improved stochastic predictions
%of such BNNs to produce better policies, even with large amounts of complex and
%high-dimensional data.
%\item In particular, our BNNs trained with $\alpha=0.5$ and together with Algorithm
%\ref{algo2} form a robust black-box policy search method that performs well
%in a wide range of scenarios.
%\end{enumerate}
%
See the appendix \ref{methods} for a short introduction to all methods we
compare to and  appendix \ref{hyperp} for the hyper-parameters used.

%\vspace{-0.25cm}
\subsection{Wet-Chicken benchmark}
\begin{figure}[t]
\centering
\begin{subfigure}{.32\textwidth}
\includegraphics[width=\textwidth]{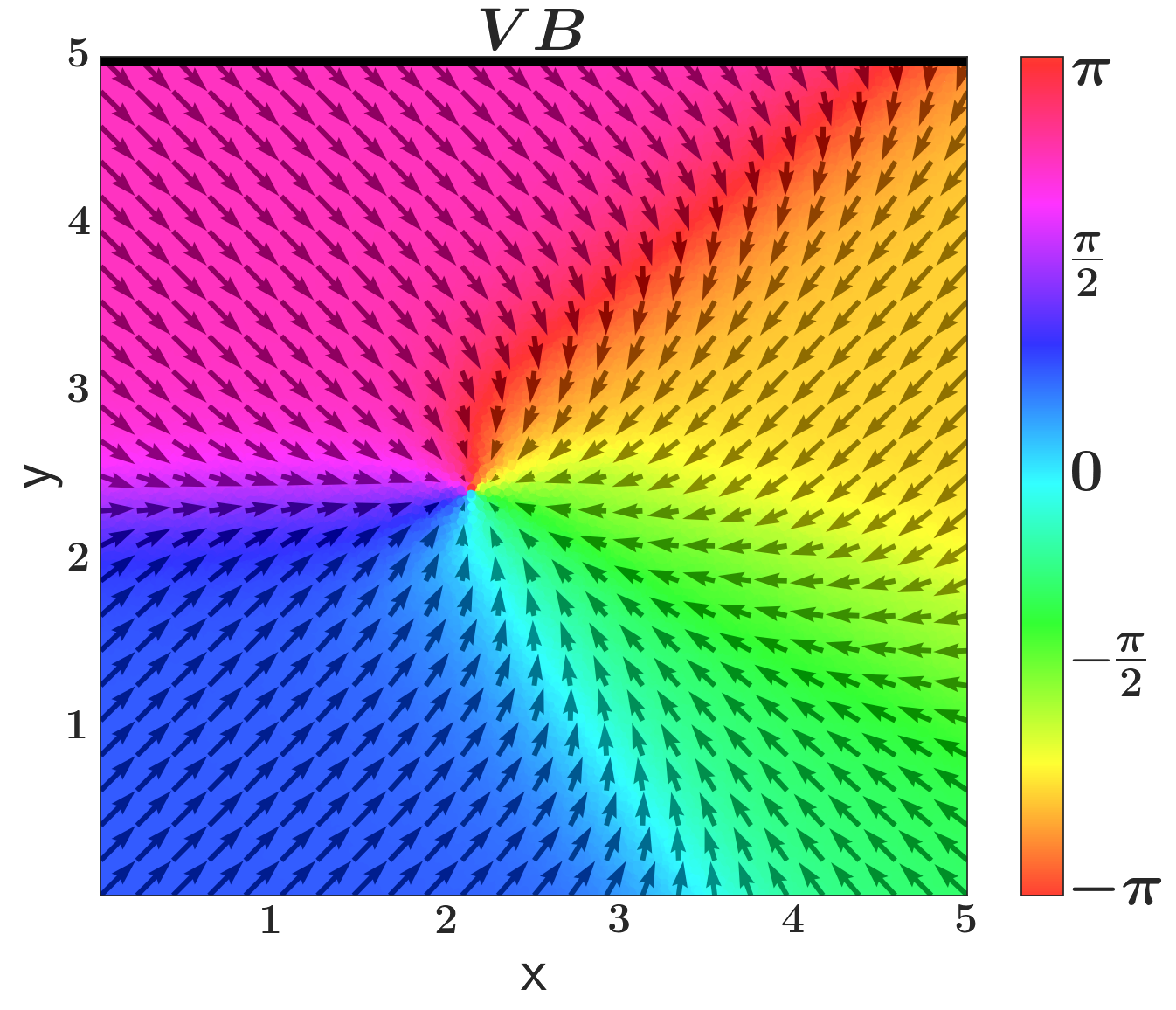}
\refstepcounter{subfigure}\label{fig:wcpol1}
\end{subfigure}
%\begin{subfigure}{.245\textwidth}
%\includegraphics[width=\textwidth]{figures/WetChicken/a05_pol1b.png}
%\refstepcounter{subfigure}\label{fig:wcpol2}
%\end{subfigure}
\begin{subfigure}{.32\textwidth}
\includegraphics[width=\textwidth]{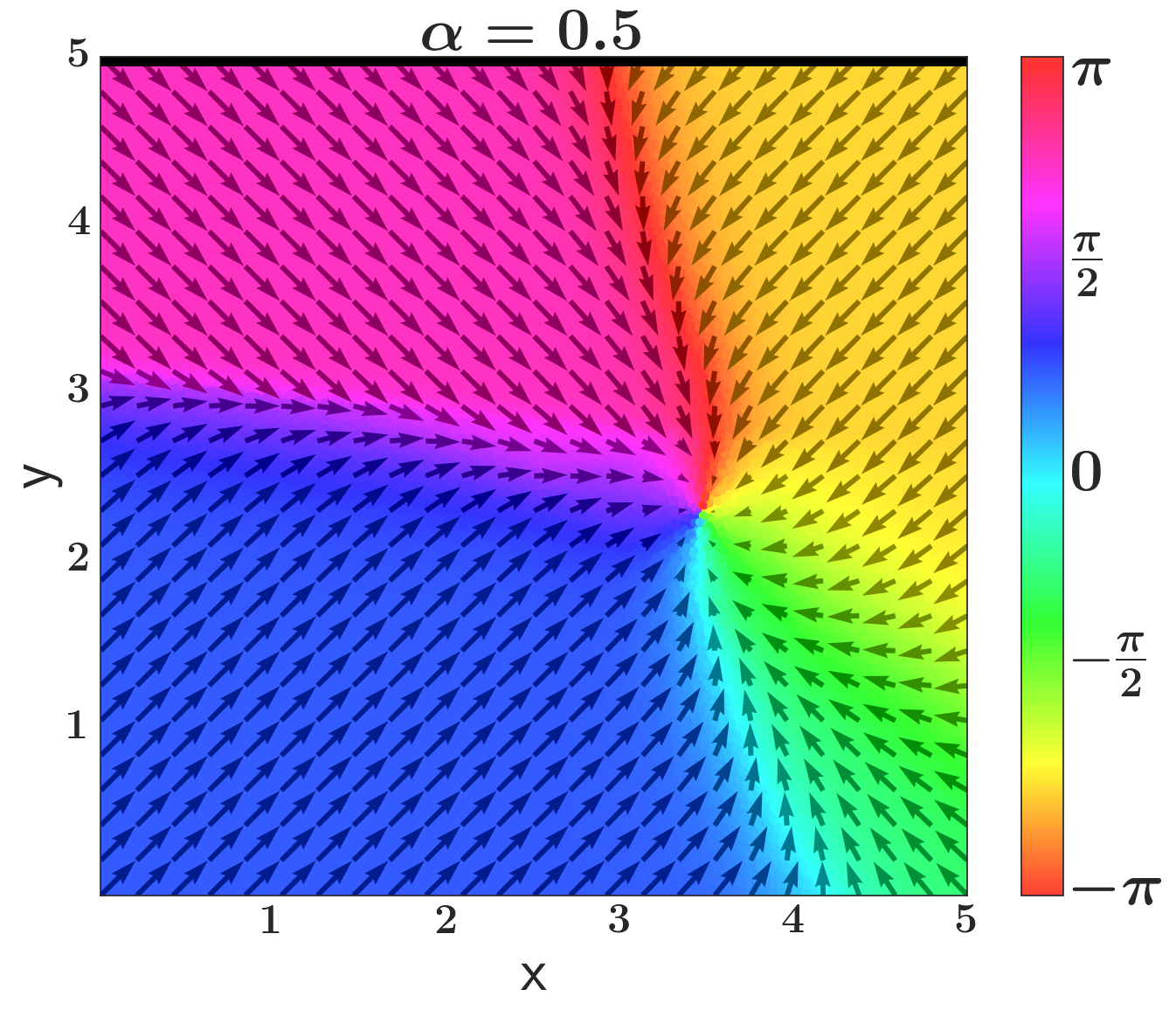}
\refstepcounter{subfigure}\label{fig:wcpol2}
\end{subfigure}
\begin{subfigure}{.32\textwidth}
\includegraphics[width=\textwidth]{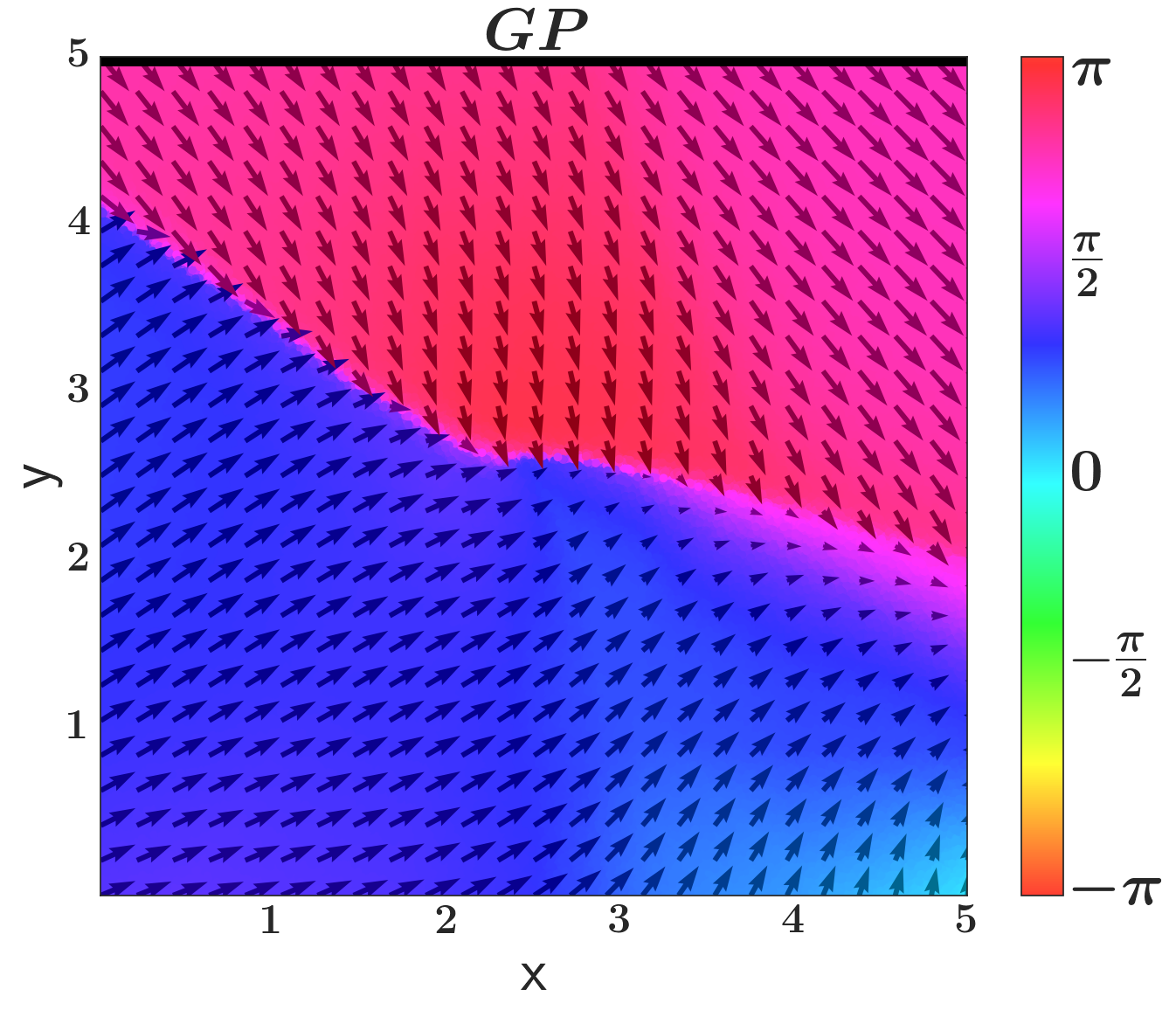}
\refstepcounter{subfigure}\label{fig:wcpol3}
\end{subfigure}
%\begin{subfigure}{.245\textwidth}
%\includegraphics[width=\textwidth]{figures/WetChicken/a00_pol1b.png}
%\refstepcounter{subfigure}\label{fig:wcpol20}
%\end{subfigure}
%\vspace{-0.5cm}
\caption{Visualization of three policies in state space. Waterfall is indicated
by top black bar. Left: policy $\pi_{VB}$ obtained with a BNN trained
with VB. Avg. reward is $-2.53$. Middle: policy $\pi_{\alpha=0.5}$ obtained with a BNN
trained with $\alpha=0.5$. Avg. reward is $-2.31$. Right:
policy $\pi_{GP}$ obtained by using a Gaussian process model. Avg.
reward is $-2.94$.  Color and arrow indicate direction of paddling of policy
when in state $\mathbf{s}_t$, arrow  length indicates action magnitude.  Best
viewed in color.} \label{fig:wc_pol}
\end{figure}

The Wet-Chicken benchmark \citep{trespwc} is  a challenging
problem for model-based policy search that presents both bi-modal and
heteroskedastic transition dynamics. We use the two-dimensional version
of the problem \citep{hans2009efficient} and extend it to the continuous case.
\newcommand{\ica}{\hspace{0.25cm}}
\renewcommand{\arraystretch}{0.94}
\begin{table*}[t]
\centering
\resizebox{\columnwidth}{!}{
\begin{tabular}{l@{\ica}r@{$\pm$}l@{\ica}r@{$\pm$}l@{\ica}r@{$\pm$}l@{\ica}r@{$\pm$}l@{\ica}r@{$\pm$}l@{\ica}|r@{\ica}}\hline
\bf{Dataset}&\multicolumn{2}{c}{\bf{ MLP
}}&\multicolumn{2}{c}{\bf{ VB 
}}&\multicolumn{2}{c}{\bf{ $\boldsymbol{\alpha{=}0.5}$
}}&\multicolumn{2}{c}{\bf{ $\boldsymbol{\alpha{=}1.0}$ 
}}&\multicolumn{2}{c}{\bf{ GP
}}&\bf{ PSO-P
}\\
\hline
Wetchicken&-2.71&0.09&-2.67&0.10&\bf{-2.37}&\bf{0.01}&-2.42&0.01&-3.05&0.06&-2.34 \\
Turbine&-0.65&0.14&-0.45&0.02&\bf{-0.41}&\bf{0.03}&-0.55&0.08&-0.64&0.18&NA \\
Industrial&-183.5&4.1&-180.2&0.6&-174.2&1.1&\bf{-171.1}&\bf{2.1}&-285.2&20.5&-145.5\\
%WetChicken
\hline
\bf{Avg. Rank}&3.6&0.3&3.1&0.2&\bf{1.5}&\bf{0.2}&2.3&0.3&4.5&0.3&\\
%0.0730 &  2.8450  &  0.0720\\
\hline
\end{tabular}}
\caption{Policy performances over different benchmarks. Printed are average values over $5$ runs with respective standard errors. Bottom row is the average rank over all $5\times3$ runs.}
\label{tab:results_pol}
%}
%\vspace{-0.25cm}
\end{table*}

\begin{table*}[t]
\centering
\resizebox{\columnwidth}{!}{
\begin{tabular}{l@{\ica}r@{$\pm$}l@{\ica}r@{$\pm$}l@{\ica}r@{$\pm$}l@{\ica}r@{$\pm$}l@{\ica}r@{$\pm$}l@{\ica}}\hline
\bf{Dataset}&\multicolumn{2}{c}{\bf{ MLP
}}&\multicolumn{2}{c}{\bf{ VB
}}&\multicolumn{2}{c}{\bf{ $\boldsymbol{\alpha{=}0.5}$
}}&\multicolumn{2}{c}{\bf{ $\boldsymbol{\alpha{=}1.0}$ 
}}&\multicolumn{2}{c}{\bf{ GP
}}\\
\hline
\bf{MSE} & \multicolumn{2}{c}{} &  \multicolumn{2}{c}{}  & \multicolumn{2}{c}{} &  \multicolumn{2}{c}{}  &  \multicolumn{2}{c}{}   \\
WetChicken&\bf{1.289}&\bf{0.013}&1.347&0.015&1.347&0.008&1.359&0.017&1.359&0.017\\
Turbine&\bf{0.16}&\bf{0.001}&0.21&0.003&0.192&0.002&0.237&0.004&0.492&0.026 \\
Industrial&0.0186&0.0052&0.0182&0.0052&\bf{0.017}&\bf{0.0046}&0.0171&0.0047&0.0233&0.0049\\
\bf{Avg. Rank}&\bf{2.0}&\bf{0.34}&3.1&0.24&2.4&0.23&2.9&0.36&4.6&0.23\\

\hline
\bf{Log-Likelihood} & \multicolumn{2}{c}{} &  \multicolumn{2}{c}{}  & \multicolumn{2}{c}{}  & \multicolumn{2}{c}{} &  \multicolumn{2}{c}{}   \\
WetChicken&-1.755&0.003&-1.140&0.033&\bf{-1.057}&\bf{0.014}&-1.070&0.011&-1.722&0.011 \\
Turbine&-0.868&0.007&-0.775&0.004&\bf{-0.746}&\bf{0.013}&-0.774&0.015&-2.663&0.131\\
Industrial&0.767&0.047&1.132&0.064&\bf{1.328}&\bf{0.108}&1.326&0.098&0.724&0.04 \\
\bf{Avg. Rank}&4.3&0.12&2.6&0.16&\bf{1.3}&\bf{0.15}&2.1&0.18&4.7&0.12\\
%0.0730 &  2.8450  &  0.0720\\
\hline
\end{tabular}}
\caption{Model test error and test log-likelihood for different benchmarks. Printed are average values over 5 runs with respective standard errors. Bottom row is the average rank over all $5\times3$ runs.}
\label{tab:results_mod}
%}
\vspace{-0.25cm}
\end{table*}
In this problem, a canoeist is paddling on a two-dimensional river. The
canoeist's position at time $t$ is $(x_t,y_t)$. The river has width $w=5$ and
length $l=5$ with a waterfall at the end, that is, at $y_t=l$. The canoeist
wants to move as close to the waterfall as possible because at time $t$ he gets
reward $r_t = -(l - y_t)$. However, going beyond the waterfall boundary makes the
canoeist fall down, having to start back again at the origin $(0,0)$. At
time $t$ the canoeist can choose an action $(a_{t,x},a_{t,y}) \in [-1,1]^2$ that represents
the direction and magnitude of his paddling. The river dynamics have stochastic
turbulences $s_t$ and drift $v_t$ that depend on the canoeist's position on the
$x$ axis. The larger $x_t$, the larger the drift and the smaller $x_t$, the
larger the turbulences.
%The underlying dynamics are heteroskedastic because
%the magnitude of the turbulences depends on $x_t$. They also present
%bi-modality because when the canoeist is close to the waterfall, the change in
%the current location can be large, if the canoeist falls down the waterfall and
%starts again at $(0,0)$, or small, if the canoeist does not fall.
The underlying dynamics are given by the following system of equations. The
drift and the turbulence magnitude are given by $v_t=3x_tw^{-1}$ and $s_t = 3.5
- v_t$, respectively. The new location $(x_{t+1},y_{t+1})$ is given
by the current location $(x_t,y_t)$ and current action $(a_{t,x},a_{t,y})$ using
\begin{align}
x_{t+1} &= \begin{cases} 
0 & \text{if } \quad x_t + a_{t,x} < 0 \\
0 & \text{if } \quad \hat{y}_{t+1} > l \\
w & \text{if }  \quad x_t + a_{t,x} > w \\
x_t + a_{t,x} & \text{otherwise}  \end{cases}\,,   
&
y_{t+1} &= \begin{cases}
0 & \text{if } \quad \hat{y}_{t+1} < 0 \\
0 & \text{if } \quad \hat{y}_{t+1} > l \\
\hat{y}_{t+1} & \text{otherwise} \end{cases}\,,
\end{align}
where $\hat{y}_{t+1} =  y_t + (a_{t,y} - 1) + v_t + s_t \tau_t$
and $\tau_t \sim \text{Unif}([-1,1])$ is a random variable that represents the
current turbulence. These dynamics result in rich transition distributions
depending on the position as illustrated by the plots in Figure
\ref{fig:wc_model}. As the canoeist moves closer to the waterfall, the
distribution for the next state becomes increasingly bi-modal (see Figure \ref{fig:wc13}) because when he is close to the waterfall, the change in
the current location can be large if the canoeist falls down the waterfall and
starts again at $(0,0)$. The distribution may also be truncated uniform for states close to
the borders (see Figure \ref{fig:wc14}).  Furthermore the system has
heteroskedastic noise, the smaller
the value of $x_t$ the higher the noise variance (compare Figure \ref{fig:wc11} with \ref{fig:wc12}). Because of these properties,
the Wet-Chicken problem is especially difficult for model-based reinforcement
learning methods. To our knowledge it has only been solved using model-free
approaches after a discretization of the state and action sets
\citep{hans2009efficient}. For model training we use a batch 2500 random state transitions.

The predictive distributions of different models for $y_{t+1}$ are shown in
Figure \ref{fig:wc_model} for specific choices of $(x_t,y_t)$ and
$(a_{x,t},a_{y,t})$. These plots show that BNNs with $\alpha=0.5$ are very
close to the ground-truth. While it is expected that Gaussian processes fail to
model multi-modalities in Figure \ref{fig:wc13}, the FTIC approximation allows
them to model the heteroskedasticity to an extent. VB captures the stochastic
patterns on a global level, but often under or over-estimates the true
probability density in specific regions. The test-loglikelihood and test MSE in
$y$-dimension are reported in Table \ref{tab:results_mod} for all methods. 
(the transitions for $x$ are deterministic given $y$).

After fitting the models, we train policies using Algorithm \ref{algo2} with a
horizon of size $T=5$. Table \ref{tab:results_pol} shows the average reward
obtained by each method. BNNs with
$\alpha=0.5$ perform best and produce policies that are very close to the
optimal upper bound, as indicated by the performance of the particle swarm
optimization policy (PSO-P). In this problem VB seems to lack robustness and
has much larger empirical variance across experiment repetitions
than $\alpha = 0.5$ or $\alpha = 1.0$.

Figure \ref{fig:wc_pol} shows three example policies, $\pi_\text{VB}$
,$\pi_{\alpha=0.5}$ and  $\pi_\text{GP}$ (Figure
\ref{fig:wcpol1},\ref{fig:wcpol2} and \ref{fig:wcpol3}, respectively). The
policies obtained by BNNs with random inputs (VB and $\alpha = 0.5$) show a
richer selection of actions. The biggest differences
are in the middle-right regions of the plots, where the drift towards the
waterfall is large and the bi-modal transition for $y$ (missed by the GP)
is more important.

%\input{cart_pole}

%\vspace{-0.25cm}

\begin{figure}[t]
\centering
\includegraphics[width=0.31\textwidth]{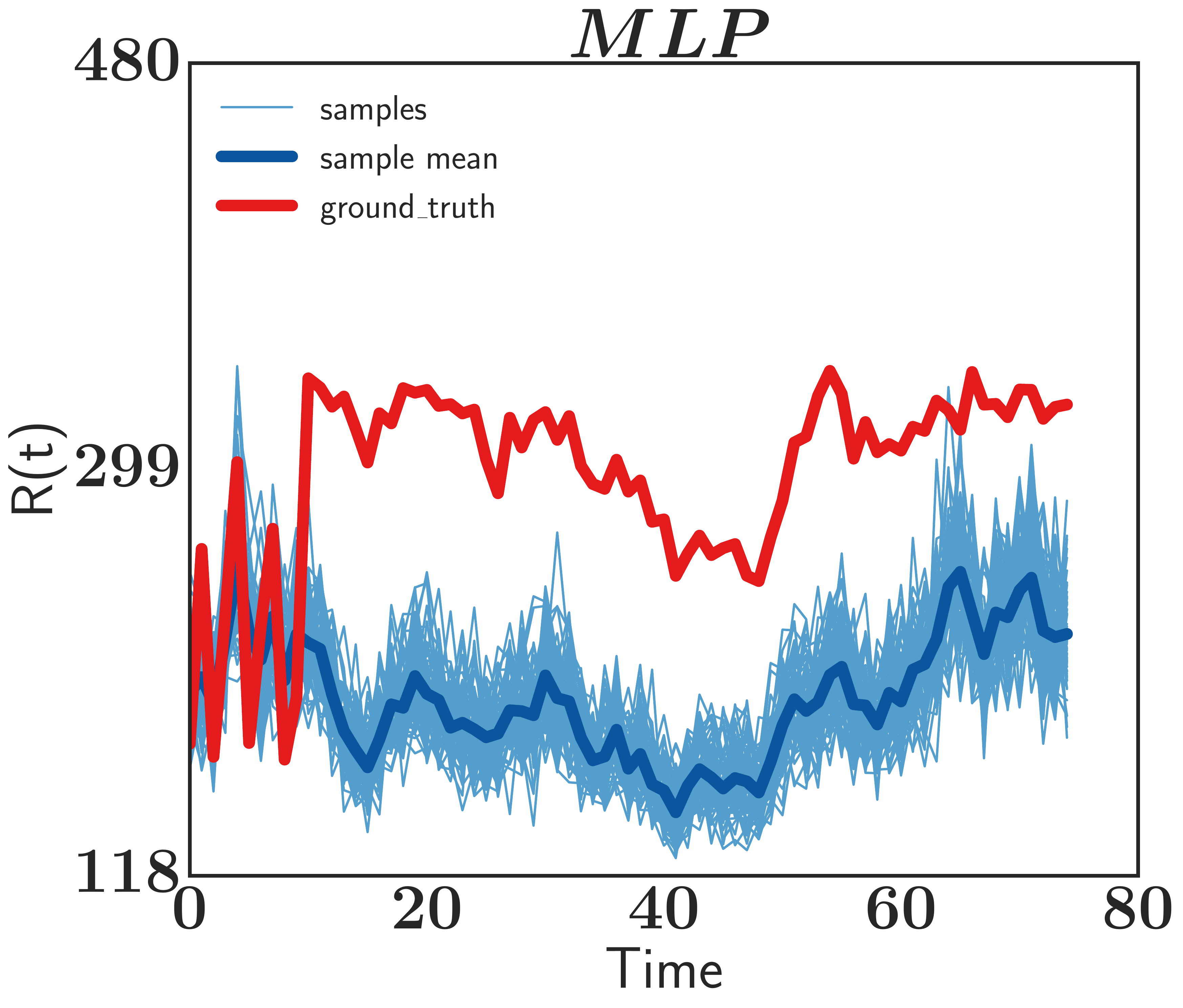}
\includegraphics[width=0.31\textwidth]{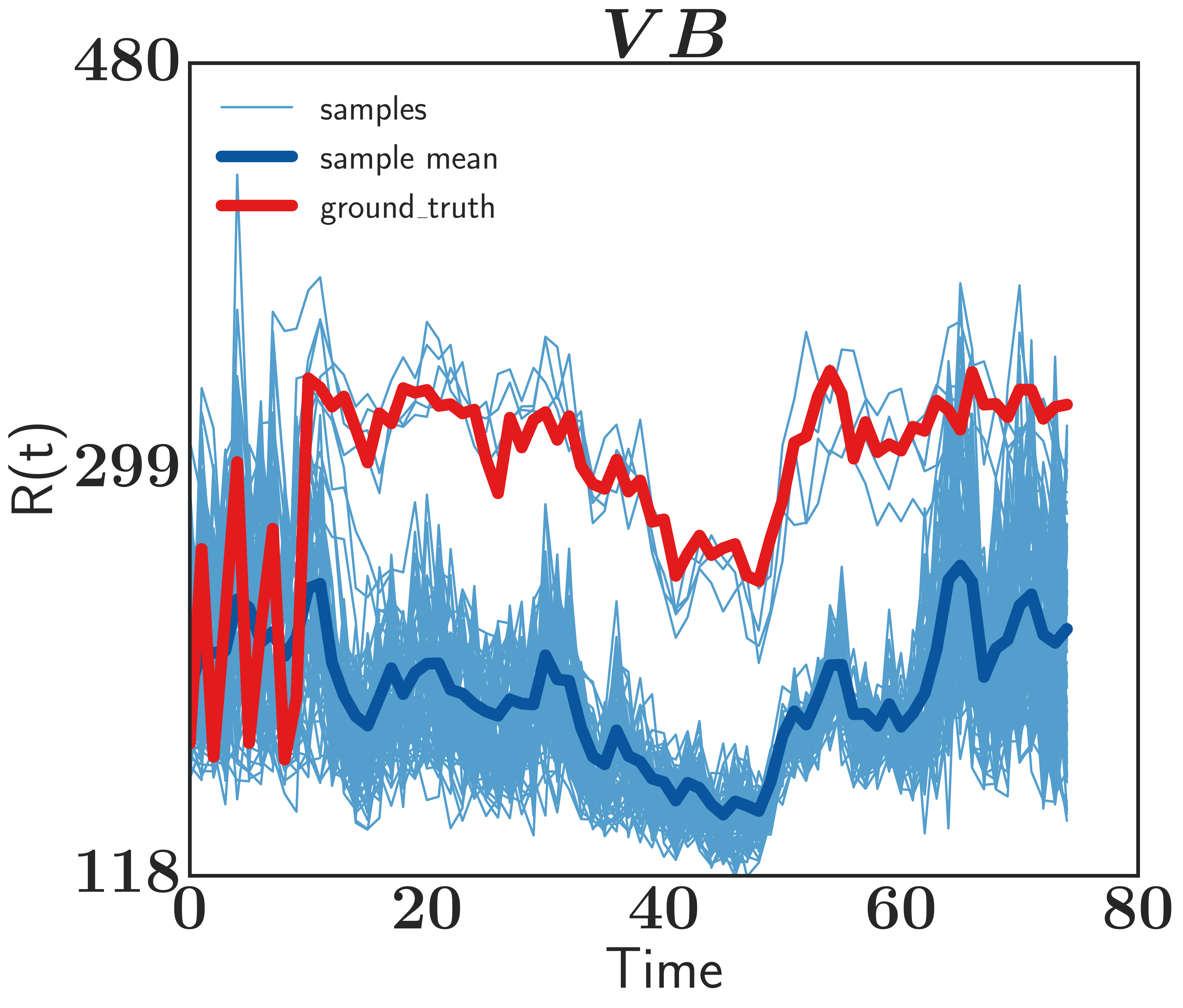}
\includegraphics[width=0.31\textwidth]{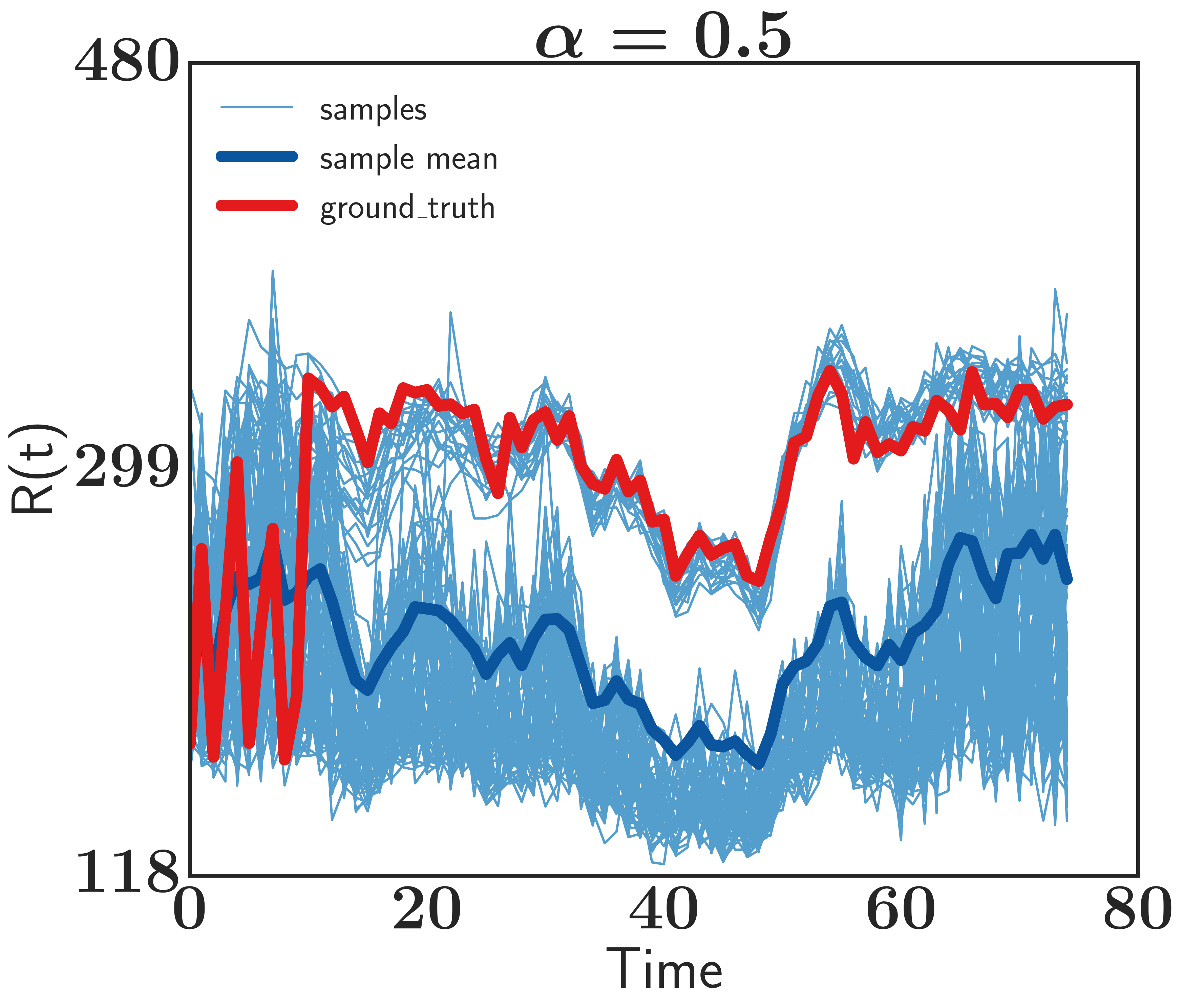}
\includegraphics[width=0.31\textwidth]{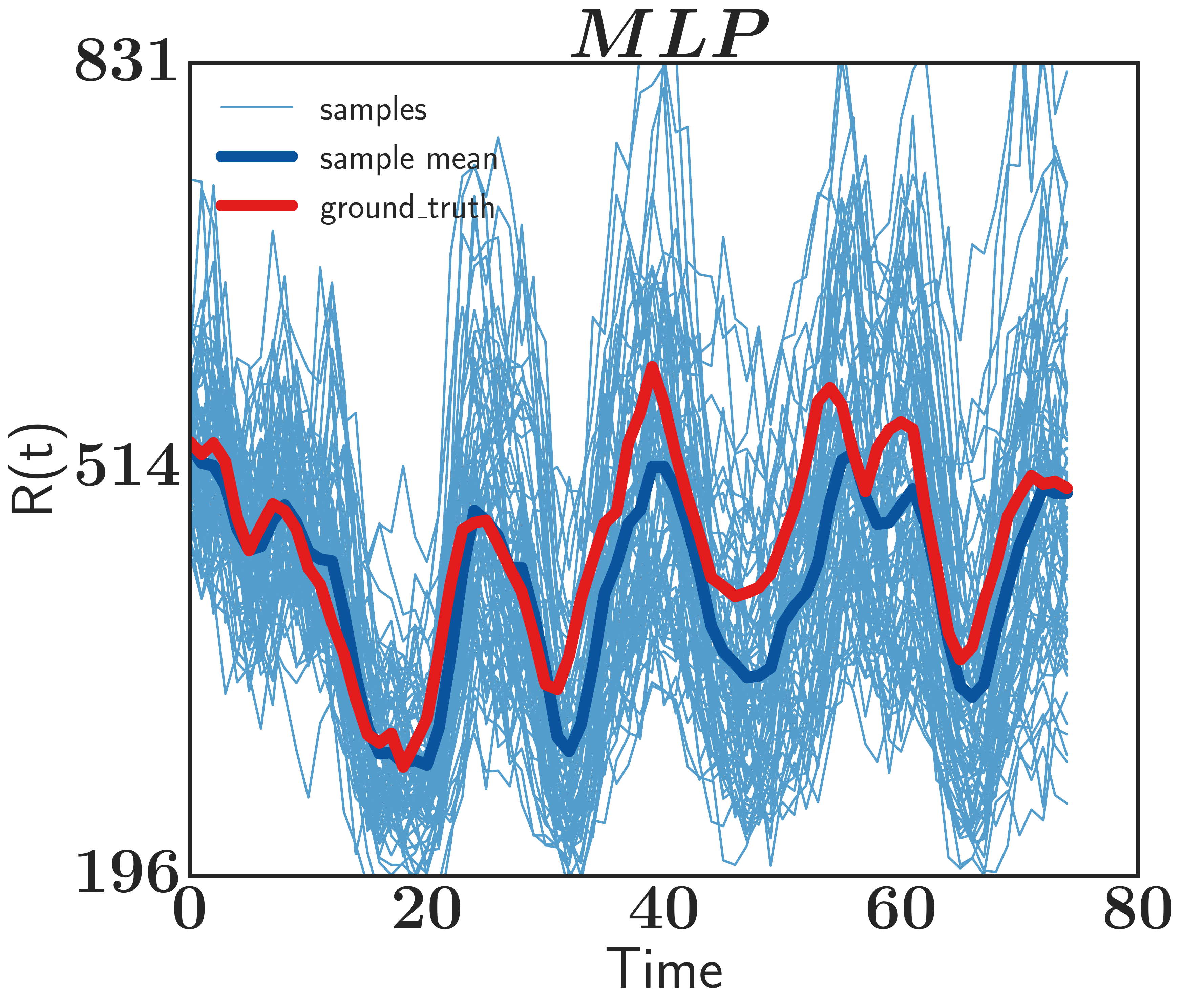}
\includegraphics[width=0.31\textwidth]{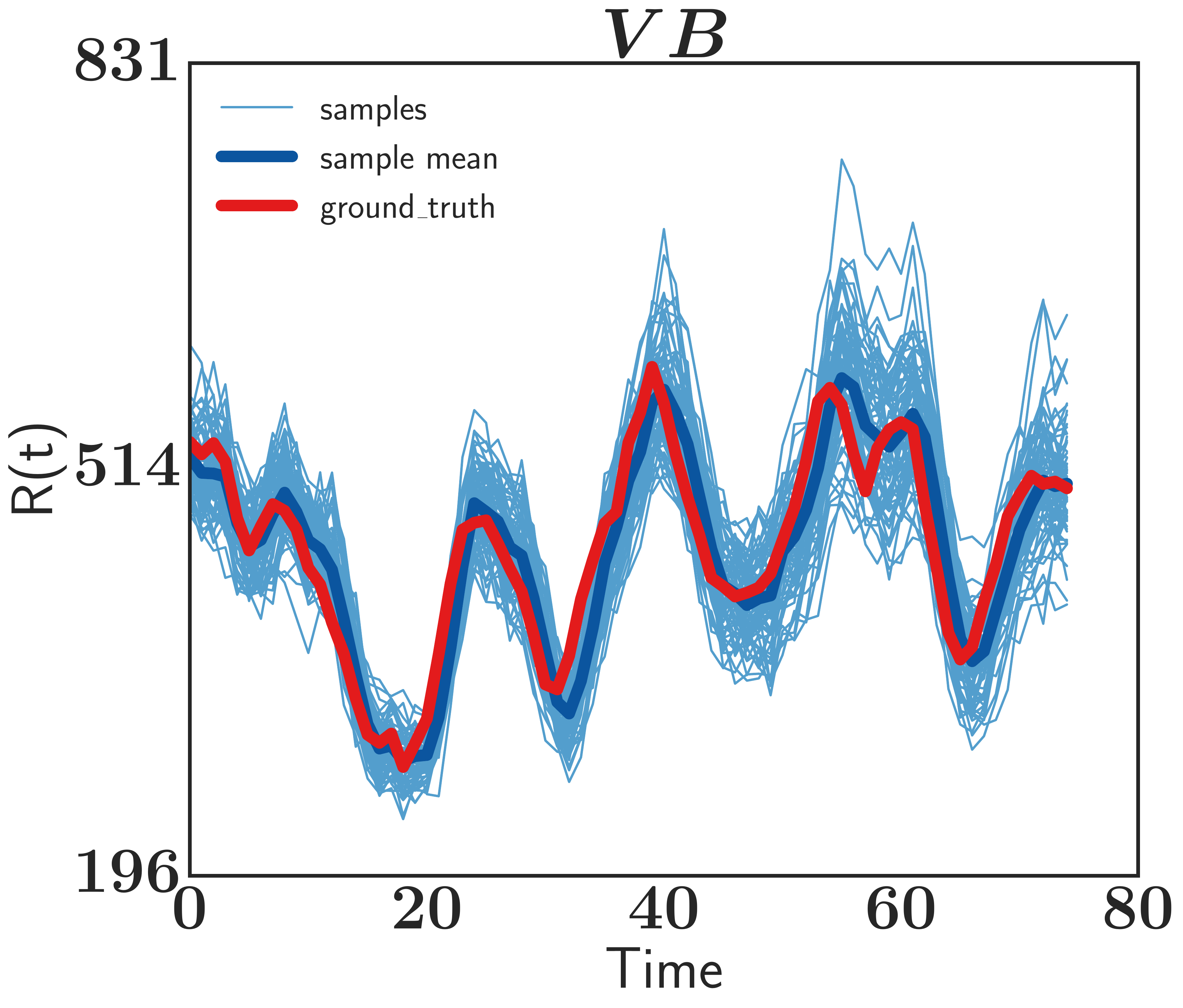}
\includegraphics[width=0.31\textwidth]{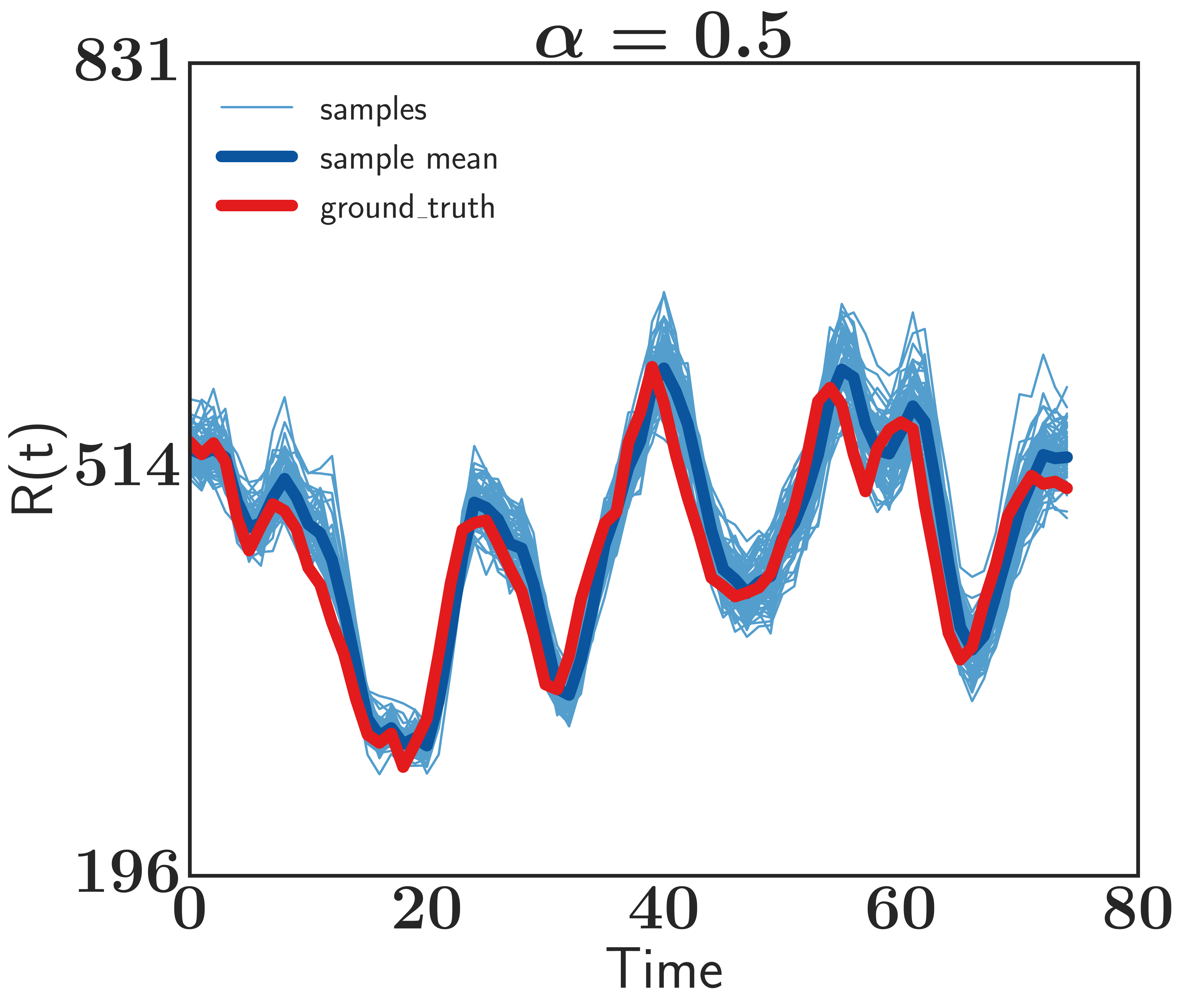}

\caption{Roll-outs of algorithm \ref{algo2} for two  starting states $\mathbf{s}_0$ (top/bottom) using different types of BNNs (left to right) with $K=75$ samples for $T=75$ steps. Action sequence $A_0,\cdots,A_{T=75}$ given by dataset for each $\mathbf{s}_0$.
From left to right: model trained using VB,$\alpha=0.5$ and $\alpha=1.0$ respectively. Red: trajectory observed in dataset, blue: sample average, light blue: individual samples.  
} 
\label{fig:turbine1}
\end{figure}

\subsection{Industrial applications} 

We now present results on two industrial cases. First,
we focus on data generated by a real gas
turbine and second, we consider a recently introduced simulator called the
\emph{"industrial benchmark"}, with code publicly
available\footnote{\url{ http://github.com/siemens/industrialbenchmark}} \citep{hein2016introduction}.
According to the authors: "The \emph{"industrial benchmark"} aims at being realistic in the sense, that it includes a variety of aspects
that we found to be vital in industrial applications."
%Our goal is to
%show the applicability of our method towards real industrial scenarios, with the
%first part including experiments on real-word data and the second part
%including results on a reproducible benchmark that has similarities to
%real-world applications.

%\vspace{-0.2cm}
\subsubsection{Gas turbine data}

For the experiment with gas turbine data we simulate a task with partial observability.  To
that end we use 40,000 observations of a 30 dimensional time-series of sensor
recordings from a real gas turbine. We are also given a cost function that
evaluates the performance of the current state of the turbine. The features in
the time-series are grouped into three different sets: a set of environmental variables
$E_t$ (e.g. temperature and measurements from sensors in the turbine) that cannot be 
influenced by the agent, a set of
variables relevant for the cost function $N_t$ (e.g. the turbines current
pollutant emission) and a set of steering variables $A_t$ that can be
manipulated to control the turbine. 

We first train a world model as a reflection of the real turbine dynamics. To
that end we define the world model's transitions for $N_t$ to have the
functional form $N_t = f(E_{t-5},..,E_{t},A_{t-5},..A_t)$. The world model
assumes constant transitions for the environmental variables: $E_{t+1} = E_{t}$.
To make fair comparisons, our world model is given by a non-Bayesian neural
network with deterministic weights and with additive Gaussian output noise.

We then use the world model to generate an artificial batch of data for
training the different methods. The inputs in this batch are still the same as in the
original turbine data, but the outputs are now sampled from the world model.
After generating the artificial data, we only keep a small subset of the
original inputs to the world model. The aim of this experiment is to learn
policies that are robust to noise in the dynamics. This noise would originate
from latent factors that cannot be controlled, such as the missing features
that were originally used to generate the outputs by the world model but which are no
longer available. After training the models for the dynamics, we use algorithm
\ref{algo2} for policy optimization. The resulting policies are then finally
evaluated in the world model.

Tables \ref{tab:results_mod}  and \ref{tab:results_pol} show the respective
model and policy performances for each method. The experiment was repeated $5$
times and we report average results.  We observe that $\alpha=0.5$ performs
best in this scenario, having the highest test log-likelihood and best policy
performance.  
%We believe that, in
%this problem, VB and $\alpha = 1.0$ either lack flexibility to model the
%stochasticity given by the partial observability of the task (VB), or have excessive
%predictive error ($\alpha = 1.0$). 

\vspace{-0.2cm}
\subsubsection{Industrial benchmark}
In this benchmark the hidden Markov state space $\mathbf{s}_t$ consists of $27$
variables, whereas the observable state $\mathbf{o}_t$ is only 5 dimensional.
This observable state consists of 3 adjustable steering variables $A_t$: 
the velocity $v(t)$, the gain $g(t)$ and the shift $s(t)$. We also observe the
fatigue $f(t)$ and consumption $c(t)$ that together form the reward signal $R(t) = -(3f(t) + c(t))$. Also visible is the setpoint $S$, a constant hyper-parameter of the benchmark that indicates the complexity of the dynamics. 
 
For each setpoint $S \in \{10,20,\cdots,100\}$ we generate $7$ trajectories of length $1000$ using random exploration. This batch with $70,000$ state transitions forms the training set. 
We use $30,000$ state transitions, consisting of $3$ trajectories for each setpoint, as test set.

For data preprocessing, in addition to the standard normalization process, we
apply a log transformation to the reward variable. Because the reward is
bounded in the interval $[0,R_{max}]$, we  use a logit transformation to
map this interval into the real line. We define the functional form for the
dynamics as $R_t = f(A_{t-15},\cdots,A_{t},R_{t-15},\cdots,R_{t-1})$. 

The test errors and log-likelihood are given in Table \ref{tab:results_mod}. We
see that BNNs with $\alpha = 0.5$ and $\alpha = 1.0$ perform best here, whereas
Gaussian processes or the MLP obtain rather poor results.

Each row in Figure \ref{fig:turbine1} visualizes long term predictions of the
MLP and BNNs trained with VB and $\alpha = 0.5$ in two specific cases.  In the
top row we see that while all three methods produce wrong predictions in
expectation (compare dark blue curve to red curve). However, BNNs trained with
$VB$ and with $\alpha=0.5$ exhibit a bi-modal distribution of predicted
trajectories, with one mode following the ground-truth very closely. By
contrast, the MLP misses the upper mode completely.  The bottom row shows that
the VB and $\alpha = 0.5$ also produce more tight confident bands in other
settings. 

Next, we learn policies using the trained models. Here we use a relatively
long horizon of $T=75$ steps. Table \ref{tab:results_pol} 
shows average rewards obtained when applying the policies to the real dynamics. 
Because both benchmark and models  have an autoregressive component, 
we do an initial warm-up phase using random exploration 
before we  apply the policies to the system and start to measure rewards.

We observe that  GPs perform very poorly in this benchmark. We believe the
reason for this is the long search horizon, which makes the uncertainties in the
predictive distributions of the GPs become very large. Tighter confidence bands, as illustrated in Figure
\ref{fig:turbine1} seem to be key for learning good policies. Overall,
$\alpha = 1.0$  performs best with $\alpha=0.5$ being very close.

%\begin{tabular}{lcc}
%\toprule
%{\bf Method} & RMSE & Log-likelihood\\
%\midrule
%MLP & {\bf 0.24}  &  1.07 \\
%$\alpha = 10^{-6}$ & {\bf 0.24} & 1.32 \\
%$\alpha = 0.5$ & 0.25 & 1.89  \\
%$\alpha = 1.0 $ & 0.27 & {\bf 2.18} \\
%\bottomrule
%\end{tabular}
%\label{tab:ibs}
%\caption{Model Performance for different BNNS}
%\end{table}

%\input{comparison_gps}

\vspace{-0.15cm}
%\vspace{-0.2cm}
\section{Related work}
%\vspace{-0.2cm}
There has been relatively little attention to using Bayesian neural networks
for reinforcement learning. In \cite{Blundell2015} a Thompson sampling approach
is used for a contextual bandits problem; the focus is tackling the
exploration-exploitation trade-off, while the work in \citet{watter2015embed}
combines variational auto-encoder with stochastic optimal control for visual
data. Compared to our approach the first of these contributions focusses on the
exploration/exploitation dilemma, while the second one uses a stochastic
optimal control approach to solve the learning problem. By contrast, our work
seeks to find an optimal parameterized policy.

Policy gradient techniques are a prominent class of policy search
algorithms \citep{peters2008reinforcement}.  While model-based
approaches were often used in discrete spaces \citep{wang2003model},
model-free approaches tended to be more popular in continuous spaces
(e.g. \citet{peters2006policy}).  

Our work can be seen as a Monte-Carlo model-based policy gradient
technique in continuous stochastic systems. Similar work was done
using Gaussian processes \citep{deisenroth2011pilco} and with
recurrent neural networks \citep{schaefer2007recurrent} . The Gaussian
process approach, while restricted to a Gaussian state distribution,
allows propagating beliefs over the roll-out procedure.  More
recently \citet{gu2016continuous} augment a model-free learning
procedure with data generated from model-based roll-outs.

% FDV: May be useful to add a few more cites so it doesn't seem like
% we've only read PILCO (as that is repeatedly referenced in the paper
% -- e.g.Model-Based Reinforcement Learning in Continuous Environments
% Using Real-Time Constrained Optimization / If you need a few more
% policy search citations http://www.ias.tu-darmstadt.de/uploads/Publications/Kober_IJRR_2013.pdf has a table 

\iffalse I'M NOT SURE IF THE REFERENCES TO
OTHER INFERENCE TECHNIQUES ARE AS RELEVANT?  MAYBE?  Recently bayesian
approaches toward neural network have seen increasing interests.Methods are
variational
inference(VI)\cite{Blundell2015,louizos2016structured},
Markov chain Monte-Carlo
approaches\cite{welling2011bayesian,korattikara2015bayesian} and Expectation
Propagation(EP) \cite{hernandez2015probabilistic,li2015stochastic}.  Also it
has been shown, that dropout\cite{srivastava2014dropout}, a recently introduced
regularisation technique, is a variational approximation of a deep Gaussian
process\cite{gal2015dropout}.  \fi

%\vspace{-0.2cm}
%\vspace{-0.1cm}
\section{Conclusion and future work}
%\vspace{-0.1cm}

We have extended the standard Bayesian neural network (BNN) model with the
addition of a random input noise source $z$. This enables principled Bayesian
inference over complex stochastic functions. We have shown that our BNNs with
random inputs can be trained with high accuracy by minimizing
$\alpha$-divergences, with $\alpha = 0.5$, which often produces better results
than variational Bayes. We have also presented an algorithm that uses random
roll-outs and stochastic optimization for learning a parameterized policy in a
batch scenario. This algorithm particular suited for industry domains.

Our BNNs with random inputs have allowed us to solve a challenging benchmark problem where
model-based approaches usually fail. They have also shown promising results 
on industry benchmarks including real-world data from a gas turbine. 
In particular, our experiments indicate that a BNN trained with $\alpha=0.5$ 
as divergence measure in conjunction with the presented algorithm for policy optimization
is a powerful black-box tool for policy search.

As future work we will consider safety and exploration. For safety, we believe
having uncertainty over the underlaying stochastic functions will allows us to
optimize policies by focusing on worst case results instead of on average
performance. For exploration, having uncertainty on the stochastic functions
will be useful for efficient data collection.

\subsection*{Acknowledgements}

Jos\'e Miguel Hern\'andez-Lobato acknowledges support from the Rafael del Pino
Foundation. The authors would like to thank Ryan P. Adams, 
Hans-Georg Zimmermann,  Matthew J. Johnson, David Duvenaud and Justin Bayer for helpful
discussions.

\bibliography{references}
\newpage
\appendix 

\section{Robustness of $\alpha = 0.5$ and $\alpha = 1.0$ when $q(\mathbf{z})$ is not learned}\label{sec:toy_problems} 

We evaluate the accuracy of the predictive distributions generated by BNNs with stochastic inputs trained
by minimizing (\ref{eq:energy}) for different BNNs parameterized by $\alpha$ in two simple regression 
problems. The first one is characterized by a bimodal predictive distribution.
The second is characterized by a heteroskedastic predictive distribution. In
the latter case the magnitude of the noise in the targets changes as a function
of the input features.

\begin{figure}[t]
\centering
\includegraphics[width=0.25\textwidth]{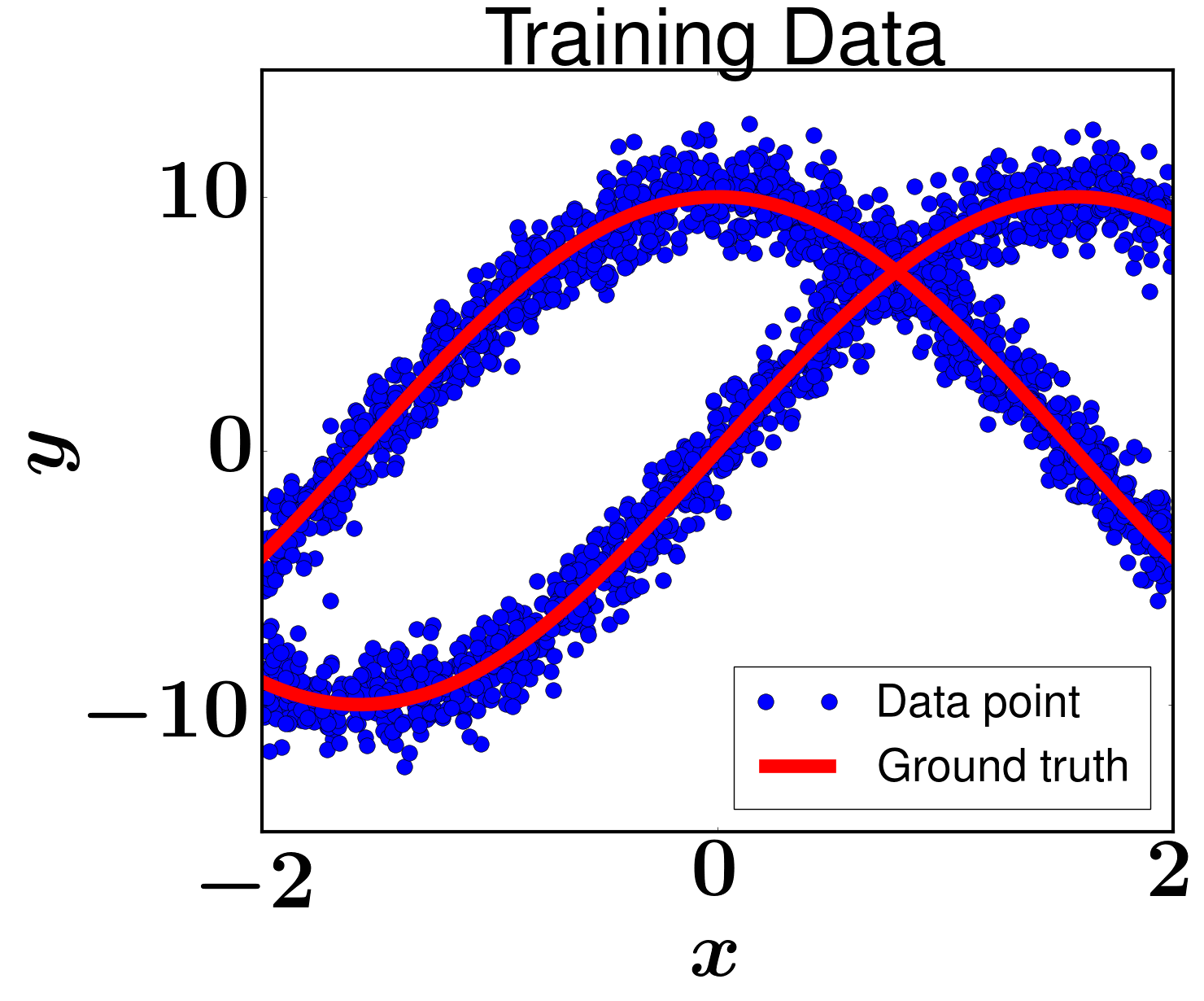}
\includegraphics[width=0.23\textwidth]{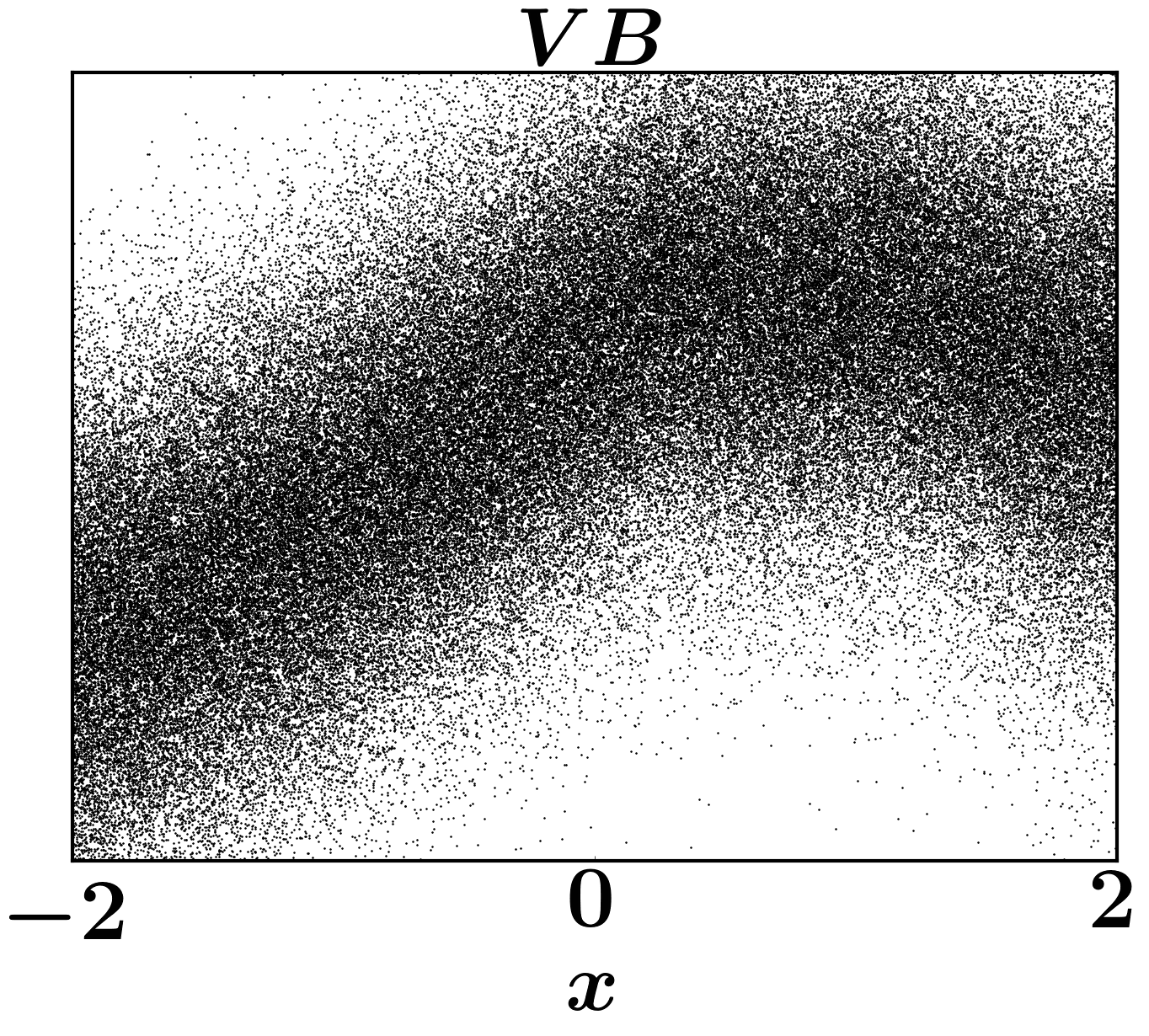}
\includegraphics[width=0.23\textwidth]{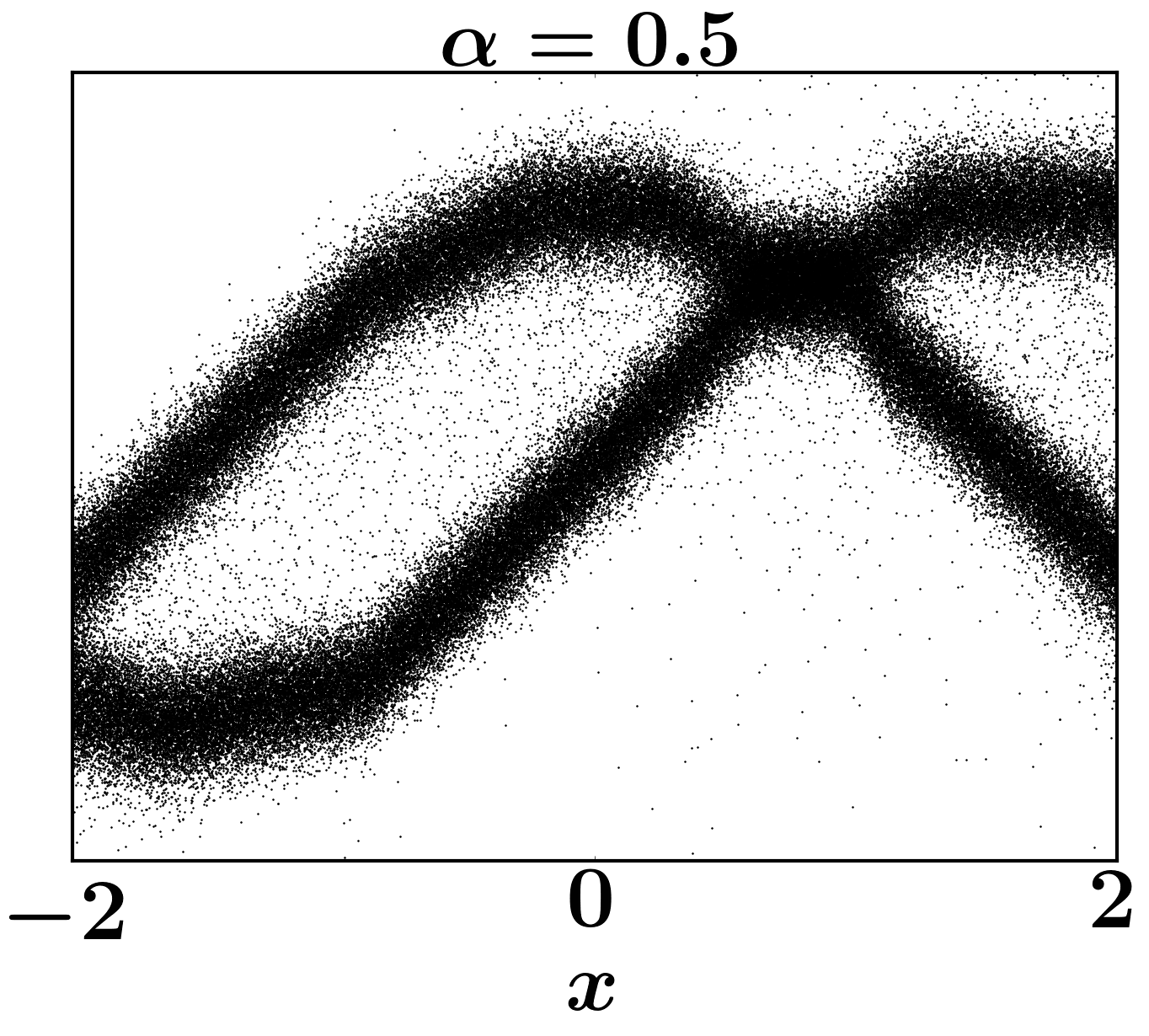}
\includegraphics[width=0.23\textwidth]{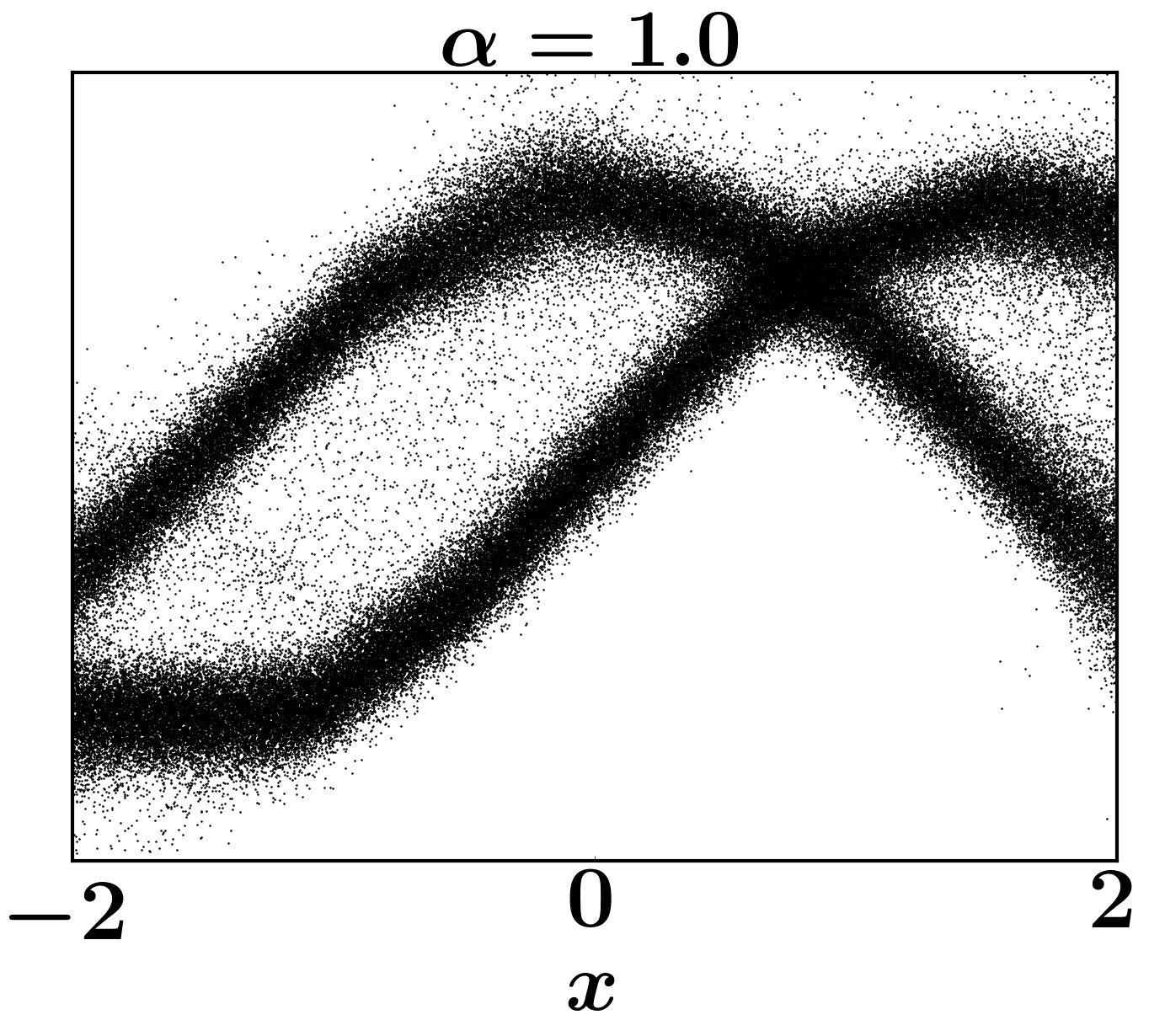}
\includegraphics[width=0.25\textwidth]{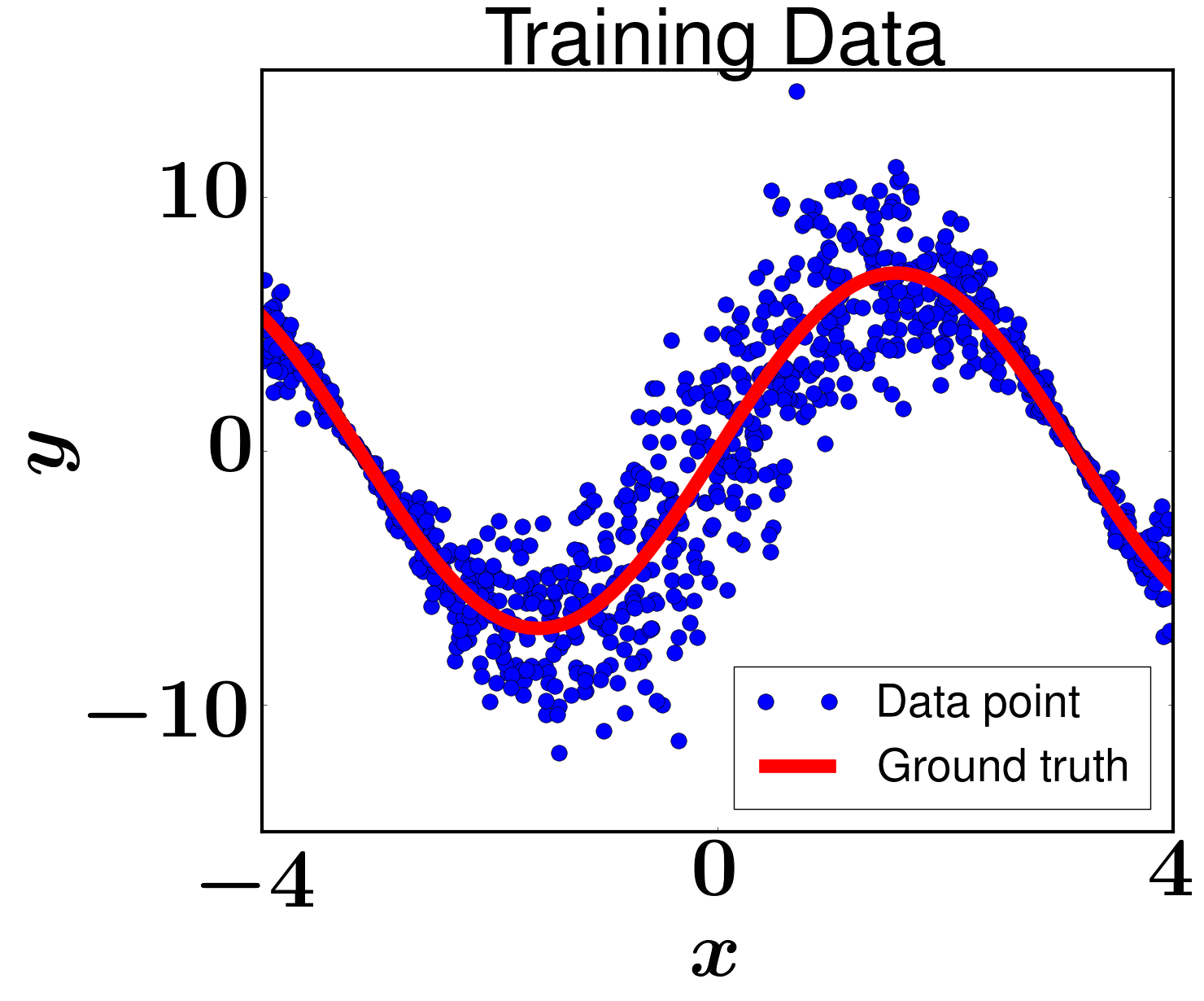}
\includegraphics[width=0.23\textwidth]{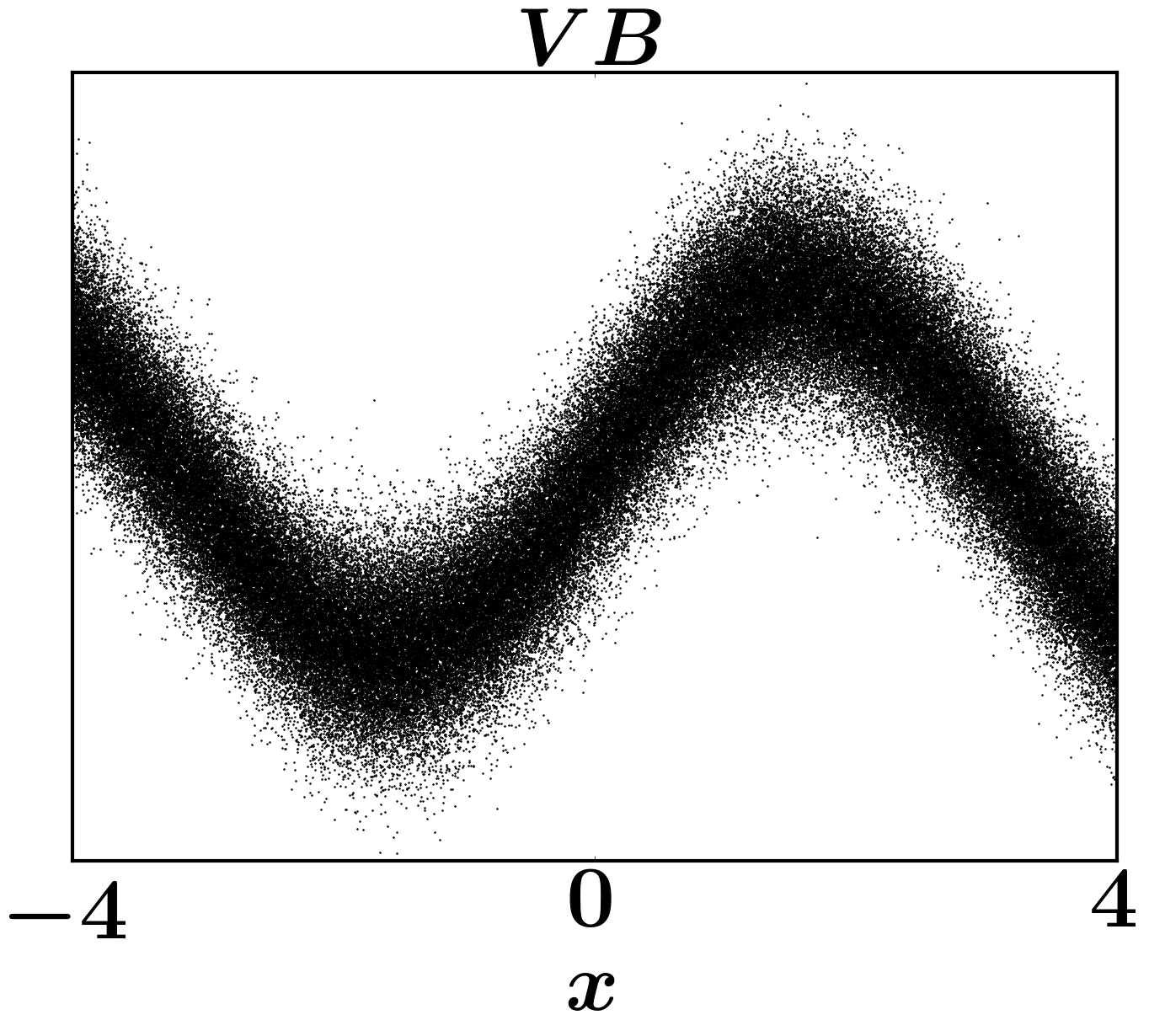}
\includegraphics[width=0.23\textwidth]{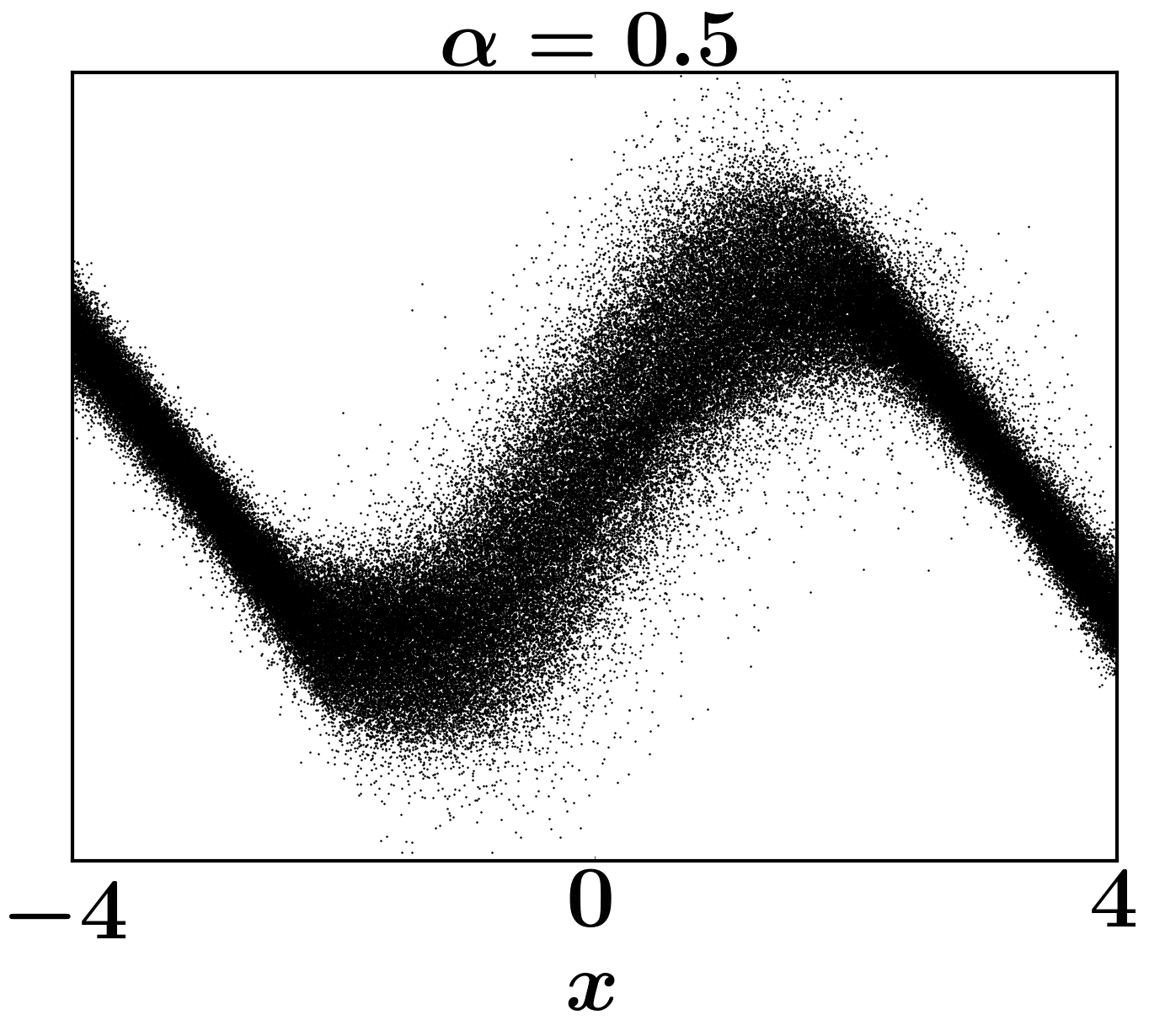}
\includegraphics[width=0.23\textwidth]{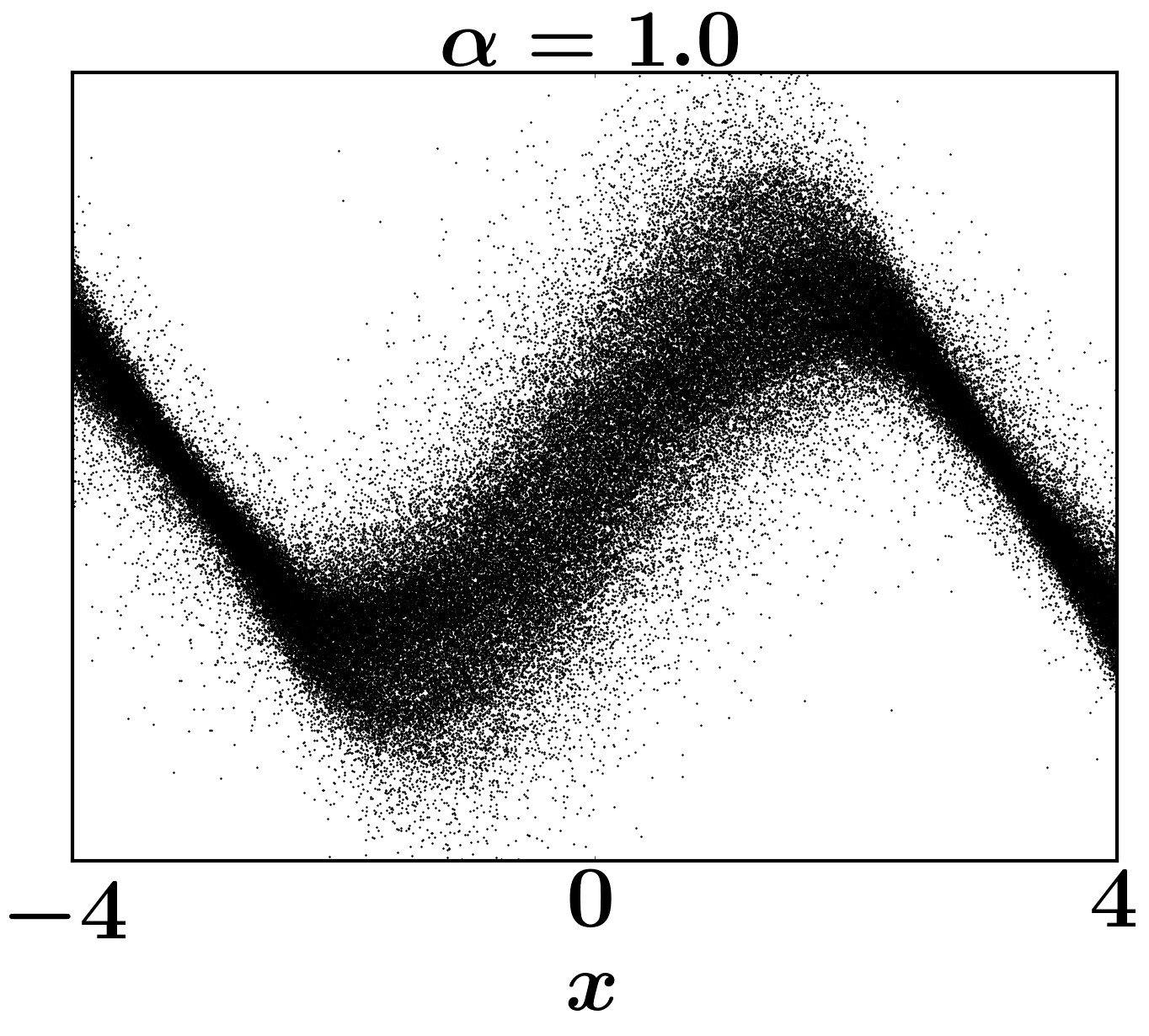}
\caption{Ground truth and predictive distributions for two toy problems introduced in main text. Top: bi-modal prediction problem, Bottom: heteroskedastic prediction problem. Left
column: Training data (blue points) and
ground truth functions (red). Columns 2-4:  predictions generated with VB, $\alpha = 0.5$ and $\alpha = 1.0$, respectively.}
\label{fig:toy_problem}
\end{figure}

In the first problem $x\in[-2, 2]$ and $y$ is obtained as $y=10\sin (x)+\epsilon$
with probability $0.5$ and $y=10\cos (x)+\epsilon$, otherwise, where $\epsilon
\sim \mathcal{N}(0,1)$ and $\epsilon$ is independent of $x$. The plot in the
top of the 1st column in Figure \ref{fig:toy_problem} shows a training dataset obtained by
sampling 2500 values of $x$ uniformly at random.  The plot clearly shows that
the distribution of $y$ for a particular $x$ is bimodal. In the second problem
$x\in[-4, 4]$ and $y$ is obtained as $y= 7 \sin (x) + 3|\cos (x / 2)| \epsilon$.
The plot in the bottom of the 1st column in Figure \ref{fig:toy_problem} shows a training
dataset obtained with 1000 values of $x$ uniformly at random. The plot clearly
shows that the distribution of $y$ is heteroskedastic, with a noise variance
that is a function of $x$.

We evaluated the predictive performance obtained by minimizing (\ref{eq:energy})
using $\alpha = 0.5$ and $\alpha = 1.0$ and also by running VB. However, we do not
learn $q(\mathbf{z})$ and keeping it instead fixed to the prior $p(\mathbf{z})$.

We fitted a
neural network with 2 hidden layers and 50 hidden units per layer using Adam
with its default parameter values, with a learning rate of 0.01 in the first
problem and 0.002 in the second problem. We used mini-batches of size 250 and
1000 training epochs. To approximate the expectations in \ref{eq:energy}, we
draw $K=50$ samples from $q$. 

The plots in the 3rd and 4th columns of Figure \ref{fig:toy_problem} show the
predictions obtained with $\alpha=0.5$ and $\alpha=1.0$, respectively. In these
cases, 
the predictive distribution is able to capture the bimodality in the
first problem and the heteroskedasticity pattern in the second problem in both cases
The
plots in the 2nd column of Figure \ref{fig:toy_problem} show the predictions
obtained with VB, which converges to suboptimal solutions in which the
predictive distribution has a single mode (in the first problem) or is
homoskedastic (in the second problem). 
Tables \ref{tab:toy_problem_bimodal} and
\ref{tab:toy_problem_heteroskedastic} show the average test RMSE and
log-likelihood obtained by each method on each problem.

These results show that Bayesian neural networks trained with $\alpha = 0.5$ or
$\alpha = 1.0$ are more robust than VB and can still model complex predictive distributions, which may
be multimodal and heteroskedastic, 
even when $q(\mathbf{z})$ is not learned and is instead kept fixed to the prior $p(\mathbf{z})$.
By contrast, VB fails to capture complex stochastic patterns in this setting.

\begin{minipage}[t]{0.45\textwidth}
\centering
%\resizebox{0.4\t}{!}{
\begin{tabular}{lcc}
\toprule
{\bf Method} & RMSE & Log-likelihood\\
\midrule
VB & {\bf 5.12} & -3.05 \\
$\alpha = 0.5$ & 5.14 & {\bf -2.10}  \\
$\alpha = 1.0$ & 5.15 & -2.11  \\
%\midrule
%VB &  5.17 & - 2.19 \\
%$\alpha=0.5$ & 5.18 & -2.11 \\
%$\alpha=1.0$  & 5.19 & -2.20 \\
\bottomrule
\end{tabular}
%}
\captionof{table}{Test error and log-likelihood for the bi-modal prediction problem. }
\label{tab:toy_problem_bimodal}
\end{minipage}
\hfill
\begin{minipage}[t]{0.45\linewidth}
\centering
%\resizebox{0.4\textwidth}{!}{
\begin{tabular}{lcc}
\toprule
{\bf Method} & RMSE & Log-likelihood  \\
\midrule
VB & {\bf 1.88} & -2.05 \\
$\alpha = 0.5$ & 1.89 & {\bf -1.78} \\
$\alpha = 1.0$ & 1.94 & -1.98 \\
%\midrule
%VB & 1.89 & -1.79 \\
%$\alpha=0.5$&-1.90&-1.80 \\
%$\alpha=1.0$&2.05&-1.95 \\
\bottomrule
\end{tabular}
%}
\captionof{table}{Test error and log-likelihood for the heteroskedastic prediction problem.}
\label{tab:toy_problem_heteroskedastic}

%\hline
%$\alpha$ & $\Gamma$ & Log-Likelihood & Reward \\ \hline
%\multirow{2}{*}{$\alpha=0$} & True & $-1.75$ & $2.37 $ \\
% & False & $-10.47$ & $2.17$ z\\ \hline
%\multirow{2}{*}{$\alpha=0.5$} & True & $-1.58$ & $2.37$ \\
% & False & $\bf{-0.99}$ & $\bf{2.65}$ \\ \hline
% \multirow{2}{*}{$\alpha=1.0$} & True & $-1.047$ & $2.46$ \\
% & False & $\bf{-0.99}$ & $\bf{2.65}$ \\ \hline
% PSO &   &  &  $2.66$ \\ 
%\hline
%\end{tabular}
\end{minipage}

\section{Methods}\label{methods}
In the experiments we compare to the following methods:

\paragraph{Standard MLP.} The standard multi-layer preceptron (MLP) is equivalent to our
BNNs, but does not have uncertainty over the weight $\mathcal{W}$ and  does not
include any stochastic inputs. We train this method using early
stopping on a subset of the training data. When we perform roll-outs using
algorithm \ref{algo2}, the predictions of the MLP are made stochastic by adding
Gaussian noise to its output. The noise variance is fixed by maximum likelihood on some
validation data after model training.

\paragraph{Variational Bayes (VB).}
The most prominent approach in training modern BNNs is to optimize the variational
lower bound \citep{Blundell2015,Houthooft2016,Gal2016}. This is in practice equivalent to
$\alpha$-divergence minimization when $\alpha \to 0$ \citep{hernandez2015black}. In our experiments we use
$\alpha$-divergence minimization with $\alpha=10^{-6}$ to implement this
method.

\paragraph{Gaussian Processes (GPs).}
Gaussian Processes have recently been used for policy search under the name of
PILCO \citep{deisenroth2011pilco}. For each dimension of the target variables,
we fit a different sparse GP using the FITC approximation
\citep{snelson2005sparse}. In particular, each sparse GP is trained using $150$ inducing
inputs by using the method stochastic expectation
propagation \citep{bui2016deep}. After this training process we approximate the
sparse GP by using a feature expansion with random basis functions (see
supplementary material of \citealt{hernandez2014predictive}). This allows us to
draw samples from the GP posterior distribution over functions, enabling the
use of Algorithm \ref{algo2} for policy training. Note that PILCO will instead
moment-match at every roll-out step as it works by propagating Gaussian
distributions. However, in our experiments we obtained better performance by
avoiding the moment matching step with the aforementioned approximation based on random basis functions.

\paragraph{Particle Swarm Optimization Policy(PSO-P).}
We use this method to estimate an upper bound for reward performance. PSO-P is a
model predictive control (MPC) method that uses the true dynamics when
applicable \citep{hein2016reinforcement}. For a given state $\mathbf{s}_t$,
the best action is selected using the standard receding horizon approach on the
real environment. Note that this is not a benchmark method to compare to, we
use it instead as an indicator of what the best possible reward can be achieved
for a fixed planning horizon $T$.

\section{Model Parameters}\label{hyperp}
For all tasks we will use a standard MLP with two hidden layer with $20$ hidden units each  as policy representation.
The activation functions for the hidden units are rectifiers:~${\varphi(x) = \max(x,0)}$.   
If present, bounding of the actions is realized using the $\tanh$ activation function on the outputs
of the policy. All models based on neural network will share the same hyperparameter. We use ADAM as learning algorithm
in all tasks.

\paragraph{WetChicken}
The neural network models are set to  2 hidden layers and 20 hidden units per
layer. We use 2500  random state
transitions for training. We found that  assuming no
observation noise by setting $\Gamma$ to a constant of $10^{-5}$ helped the
models converge to lower energy values. 

For policy training  we use a horizon of size $T=5$ and optimize the policy network for $100$ epochs,
averaging over $K=20$ samples in each gradient update, with mini-batches of
size $10$ and learning rate set to $10^{-5}$.

\paragraph{Turbine}
The world model and the BNNs have two hidden layers with
$50$ hidden units each. For policy training and world-model evaluation we
perform a roll-out with horizon $T=20$. For learning the policy we use
minibaches of size $10$ and draw $K=10$ samples from $q$.  

\paragraph{Industrial Benchmark}
For the neural network models we use two hidden layers with $75$ hidden units.We use a horizon of $T=75$, training for $500$ epochs with batches of size
$50$ and $K=25$ samples for each rollout.

\section{Computational Complexity}
\subsection*{Model Training}
All models were trained using theano and a single GPU. Training  the standard neural network is fast, the training time for this method was between 5 - 20 minutes, depending on data set size and dimensionality of the benchmark.
In theano, the computational graph of the BNNs is similar to that of an ensemble of standard neural networks. The training time for the BNNs varied between 30 minutes to 5 hours depending on data size and dimensionality of benchmark.
The sparse Gaussian Process was optimized using an expectation propagation algorithm and after training, it was approximated
with a Bayesian linear model with fixed basis functions whose weights are initialized randomly (see Appendix \ref{methods}). 
We choose the inducing points in the GPs and the number of training epochs for these models so that the resulting 
training time was comparable to that of the BNNs.
\subsection*{Policy Search}
For policy training we used a single CPU. All methods are of similar complexity as they are all trained using Algorithm 1.
Depending on the horizon, data set size and network topology, training took between 20 minutes  (Wet-Chicken, $T=5$), 3-4 hours  (Turbine, $T=20$) and 14-16  hours (industrial benchmark, $T=75$).
\bibliographystyle{iclr2017_conference}

\end{document}